%% file: LeanStereo.tex
\documentclass[conference]{IEEEtran}
\IEEEoverridecommandlockouts

\usepackage{cite}
\usepackage{amsmath,amssymb,amsfonts}
\usepackage{algorithmic}
\usepackage{graphicx}
\usepackage{textcomp}

\usepackage{xcolor}
\def\BibTeX{{\rm B\kern-.05em{\sc i\kern-.025em b}\kern-.08em
    T\kern-.1667em\lower.7ex\hbox{E}\kern-.125emX}}

\makeatletter
\DeclareRobustCommand\onedot{\futurelet\@let@token\@onedot}
\def\@onedot{\ifx\@let@token.\else.\null\fi\xspace}
\usepackage{xspace}
\include{def}

\usepackage{times}
\usepackage{epsfig}
\usepackage{graphicx}
\usepackage{amsmath}
\usepackage{amssymb}
\usepackage{booktabs}
\usepackage{subcaption}

\def\tabref#1{Table~\ref{table:#1}}
\def\figref#1{Fig.~\ref{fig:#1}}

\def\opix{$ 1 $\textit{px}\xspace}
\def\tpix{$ 2 $\textit{px}\xspace}
\def\dpix{$ 3 $\textit{px}\xspace}
\usepackage{siunitx}
\usepackage{numprint}
\usepackage{makecell}
\usepackage{multirow}
\usepackage{arydshln, booktabs}
\usepackage{siunitx}
\usepackage{subcaption}
\usepackage{enumitem}
\usepackage{rotating}
\usepackage{adjustbox}
\usepackage{xcolor}
\usepackage{soul}
\usepackage{tablefootnote}
\usepackage[absolute,overlay]{textpos}

\begin{document}

\title{\methodname: A Leaner Backbone based Stereo Network\\
\thanks{Funded by the Deutsche Forschungsgemeinschaft (DFG, German Research Foundation) under Germany’s Excellence Strategy – EXC number 2064/1 – Project number 390727645}
}

\author{\IEEEauthorblockN{Rafia Rahim}
\IEEEauthorblockA{\textit{Cognitive Systems Group} \\
\textit{University of Tuebingen}\\
Tuebingen, Germany \\
rafia.rahim@uni-tuebingen.de}
\and
\IEEEauthorblockN{Samuel Woerz}
\IEEEauthorblockA{
	\textit{University of Tuebingen}\\
	Tuebingen, Germany \\
	samuel.woerz@student.uni-tuebingen.de}
\and
\IEEEauthorblockN{Andreas Zell}
\IEEEauthorblockA{\textit{Cognitive Systems Group} \\
	\textit{University of Tuebingen}\\
	Tuebingen, Germany \\
	andreas.zell@uni-tuebingen.de}
}
\begin{textblock*}{21cm}(0.2cm, 26.7cm) 
	\noindent
	\copyright \ 2023 IEEE. Personal use of this material is permitted. Permission from IEEE must be obtained for all other uses, in any current or future media, including reprinting/republishing this material for advertising or promotional purposes, creating new collective works, for resale or redistribution to servers or lists, or reuse of any copyrighted component of this work in other works.
\end{textblock*}
\maketitle

\begin{abstract}
 Recently, end-to-end deep networks based stereo matching methods,  mainly because of their performance,  have gained popularity. However, this  improvement in performance comes at the cost of increased computational and memory bandwidth requirements, thus necessitating specialized hardware (GPUs); even then, these methods have large inference times compared to classical methods. This limits their applicability in real-world applications. Although we desire high accuracy stereo methods albeit with reasonable inference time. To this end, we propose a fast end-to-end stereo matching method. Majority of this speedup comes from integrating a leaner backbone.  To recover the performance lost because of a leaner backbone, we propose to use learned attention weights based cost volume combined with \lossname for stereo matching. Using \lossname not only improves the overall performance of the proposed network  but also leads to faster convergence. We do a detailed empirical evaluation of different design choices and show that our method requires $4 \times$ less operations and is also about $9$ to  $14\times $ faster compared to the state of the art methods like \acvnet~\cite{xu2022acvnet},  \leastereo~\cite{cheng_learning_2019} and \cfnet~\cite{shen2021cfnet} while giving comparable  performance\footnote{Code: https://github.com/cogsys-tuebingen/LeanStereo}
\end{abstract}


\section{Introduction}
\input{introduction}

\section{Related Work}
\input{related-work}

\section{Methodology}
\label{sec:methodology}
\input{methodology}
\section{Experiments and Results}
\label{sec:experiments and results}
\input{experiments-results}
\section{Conclusions}
In this work, we exploit the recent advancements in \dnn design and propose a fast stereo network with a lightweight backbone. Moreover, to recover the lost performance, due to lighter backbone, we propose to use learned attention weights for refining the cost-volume and train the network with \lossname. Our lean backbone based network trained with \lossname and attention-based cost volume: (i) remains faster and give superior performance relative to real-time and 2D methods; and (ii) gives decent performance compared to recent state of the art \td methods while being much faster (up to $ 9 \mbox{ to } 14 \times $). Based on our work we can conclude that although it is feasible to design lighter and faster \td stereo methods by leveraging the recent advancements in network designs of overlapping tasks such as image segmentation, \etc, however simple replacement will lead to deteriorated performance. To recover this performance loss, task specific calibrations like introduction of better feature merging step and loss functions are necessary.

\bibliographystyle{IEEEtran}
\bibliography{IEEEabrv,Stereo}

\end{document}

%% file: def.tex
\newif\ifwithlatentalgo
\newif\iflong
\newif\iflettrine
\newif\ifiksvm
\newif\ifaptable
\longtrue
\newif\ifmultitables
\multitablestrue
\newif\ifappendix
\appendixtrue
\newif\iflongtables
\newif\ifltpap
\newif\ifnooverview
\def\bisenet{BiSeNet\xspace}
\def\dnn{deep neural networks\xspace}
\def\psm{PSMNet\xspace}
\def\ganet{GANet\xspace}

\def\mnet{MobileNet\xspace}
\def\acvnet{ACVNet\xspace}
\def\gwcnet{GwcNet\xspace}
\def\leastereo{LEAStereo\xspace}
\def\cfnet{CFNet\xspace}

\def\stereologloss{LogL1Loss\xspace}
\def\lossname{LogL1 loss\xspace}

\def\sb{shallow branch\xspace}
\def\SB{Shallow branch}
\def\db{deep branch\xspace}
\def\DB{Deep branch}
\def\methodname{LeanStereo\xspace}

\def\macs{MACs\xspace}

\def\kitti{KITTI\xspace}
\def\kittif{KITTI\xspace2015\xspace}
\def\kittit{KITTI\xspace2012\xspace}
\def\sf{SceneFlow\xspace}

\def\td{3D\xspace}

\def\tdcs{3D convolutions\xspace}

\def\ie{\emph{i.e}\onedot}
\def\Ie{\emph{I.e}\onedot}

\def\wrt{\emph{w.r.t}\onedot}

\def\cf{\emph{c.f}\onedot}

\def\etc{\emph{etc}\onedot}

\def\eg{\emph{e.g}\onedot}

\def\etal{\emph{et al}\xspace}

\def\para#1{{\bf #1.}}

\def\mbold#1{\textbf{#1}}
\def\mitalic#1{\textit{#1}}

\newif\ifwithcomments 
\withcommentstrue
\ifwithcomments

\long\def\comment#1{\para{\sffamily\{\color{red}#1}\} }

\long\def\discuss#1{{\color{ForestGreen}\mbold{\underline{DISCUSS:}}#1} }
\long\def\diagrams#1{\\{\color{Red}\mbold{\underline{Diagram:}}#1} }

\long\def\idea#1{{\\\color{Magenta} \mbold{Idea:}#1}}
\long\def\revcomment#1{Reviewer BMVC:#1 }
\else

\long\def\comment#1{}
\long\def\discuss#1{ }
\long\def\revcomment#1{ }
\long\def\diagrams#1{ }
\long\def\idea#1{}
\fi

\def\l2norm{\mitalic{L2Norm}}

\def\figref#1{Fig.~\ref{fig:#1}}

\def\tabref#1{Table~\ref{table:#1}}
\def\secref#1{Sec.~\ref{sec:#1}}

\def\eqref#1{Eq.~(\ref{eq:#1})}

\newcounter{rno}

%% file: introduction.tex
Depth estimation from the input pair of rectified stereo images has been a long investigated computer vision task. In depth estimation the goal is to find the \textit{disparity} of each pixel from the input pair of rectified left and right images where the disparity is defined as horizontal displacement between the matching pixels. \Ie for a given pixel at position $ (x,y) $ in the left image, if the matching pixel location in right images is $ (x+d,y) $, then the disparity of the pixel $ (x,y) $ is  $d $ pixels. Once these disparities are estimated for each pixel, they are converted to depth values using triangulation.

\begin{figure}[!t]  
	\begin{center}
		\includegraphics[width=1\linewidth]{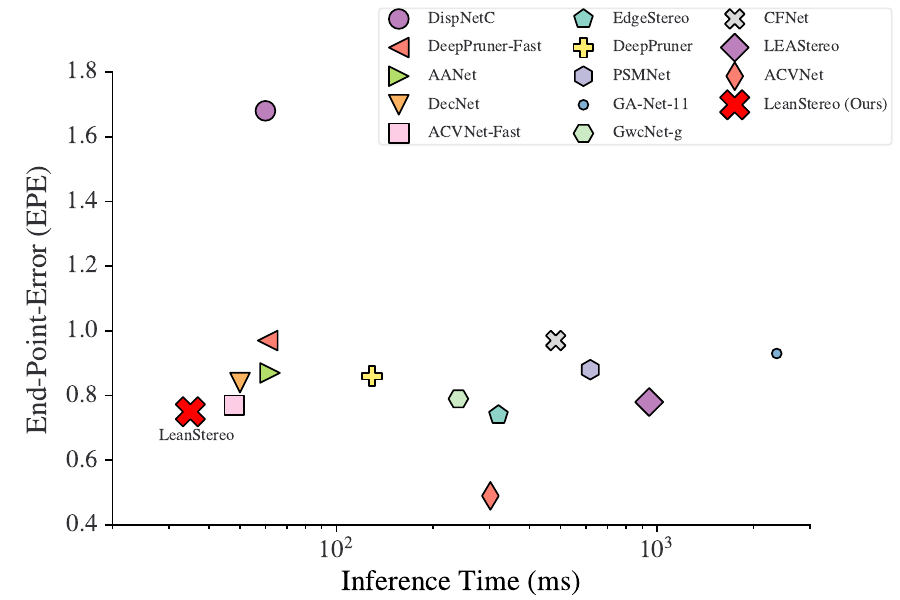}
		\vspace*{-20pt}
	\end{center}        
	\caption{Comparison of \methodname with other state of the art methods. Our proposed method is faster and has comparable performance with state of the art \td methods.}
	\label{fig:comparison}
	\vspace*{-0.2cm}
\end{figure}

Although classical stereo methods have long been optimized for the performance \cite{hosni2012fast, yoon2006adaptive,min2011revisit, scharstein2002taxonomy, poggi_synergies_2020} and been extensively used in industry applications like robotic, assisted and autonomous driving, \etc.  They still lack in performance and require complicated tuning for real-world applications \cite{hirschmuller_stereo_2008,poggi_synergies_2020}.  In comparison, recently end-to-end \dnn based methods \cite{kendall2017endtoend,chang_pyramid_2018,zhang_ga-net:_2019, xu2022acvnet} have started to give consistently superior performance. However these performant \dnn methods have huge computational requirements and are thus unfit for real-world applications -- there have been efforts to design stereo networks optimized for speed \cite{ wang_anytime_2019,khamis2018stereonet} but they lack on the performance front. Here our goal is to reduce the computational footprint of state of the art stereo \dnn methods while matching their performance. 

End-to-end \dnn based stereo methods are generally composed of four modules \cite{poggi_synergies_2020} including: (1) a backbone network, which is responsible for extracting the features from the given input stereo pair; (2) a cost volume that merges the left and right image features obtained via backbone network; (3) a cost aggregation network to process and aggregate useful information from the cost volume; and (4) a disparity regression module to regress the disparities from aggregated information. 

Broadly, these end-to-end stereo methods can be categorized into 2D methods and \td methods, depending upon the dimensions of the cost volume and convolution operations used in the cost aggregation and disparity regression modules \cite{poggi_synergies_2020}. Overall \td stereo methods perform better than 2D methods, however as shown in \cite{rahim2021separable, shamsafar2022mobilestereonet},  these performance gains in \td come at the cost of increased computational requirements and processing time. Therefore, in our work, we focus on achieving the performance of the \td stereo methods but at a fraction of their cost.

To this end, we propose, inspired by the real-time backbones from object detection \cite{sandler2018mobilenetv2,wang2018pelee} and semantic segmentation \cite{yu2021bisenet}, using a leaner backbone for \td stereo networks. This backbone contains two branches: a `\SB' responsible for capturing the fine details and a `\DB' that captures the semantic information from the input images. The motivation behind using this two branch architecture is to dedicate a part of the network to learn low-level and high-level features explicitly. 
To bridge the performance gap incurred by using a leaner backbone we 
propose to use \stereologloss combined with learned attention based cost volume.

Overall, we made the following contributions: (i) we propose to use a  two-branch based leaner backbone for \td stereo methods to capture pixel-level and semantic level information from the stereo images explicitly; (ii) we propose to use \lossname to model and improve the smaller disparity errors more explicitly; (iii) we do a detailed empirical evaluation of different design choices and show that our network is faster and gives comparable performance \wrt to other \td stereo methods. For example, our optimized network is $\sim 17\times $,  $ 9\times $, $27\times$, $14\times $ faster than \psm\cite{chang_pyramid_2018},  \acvnet \cite{xu2022acvnet}, \leastereo~\cite{cheng2020hierarchical} and \cfnet~\cite{shen2021cfnet} respectively -- \cf \figref{comparison}, note that we have reproduced all these metrics using identical settings on same hardware for most of the open source methods -- \cf \secref{inference-time}. 

%% file: related-work.tex
 
Traditionally in the computer vision domain, improving the performance of the methods had been a pivot point; nowadays, is starting to shift towards real-time methods \cite{sandler2018mobilenetv2, tan2019efficientnet, yu2021bisenet} as \dnn are becoming more and more resource/data hungry. Given this shift towards designing lightweight networks for other computer vision tasks, there is a need to explore the mechanisms to make the stereo network lightweight and real-time.

As discussed earlier stereo methods can be categorized in two categories, \ie 2D and \td methods. In 2D methods, we have a \td cost volume with dimensions $ H \times W \times C $, where $ H $ is the height, $ W $ is the width and $ C $ is the number of features/channels of the volume. Given a  \td cost volume, we need a cost aggregation and disparity regression network that contains 2D convolution operations. In contrast, \td stereo methods have a $ 4D $ cost volume of dimension $ H \times W \times D \times C $, where $ D $ denotes the disparity dimension; therefore, we need \td convolution operations to aggregate the features maps and regress disparities. 

Although 2D stereo methods \cite{mayer_large_2016, liang_learning_2018, tonioni2019real, yang_segstereo_2018, song_edgestereo_2018} are faster in comparison with \td  stereo methods, they still lack in  performance.  
\td stereo methods became popular since GCNet \cite{kendall_end--end_2017} exploited the geometrical information of stereo matching to build a costly 4D volume that improved the overall performance of the stereo networks drastically.

Since then, a number of \td stereo methods have been proposed to improve the performance, speed or both for the \td stereo networks.  \gwcnet \cite{guo_group-wise_2019}  builds on the work of \cite{kendall_end--end_2017} and introduces the group-wise correlation to build the cost volume that benefits from cost volume designs of both 2D architectures (correlation) and \td architectures (concatenation).
\psm \cite{chang_pyramid_2018} uses spatial pyramid pooling layers to increase the receptive field for better performance.

AnyNet \cite{wang_anytime_2019} exploits the coarse-to-fine resolutions strategies that had been shown successful for 2D networks to reduce the computational complexity albeit at the cost of performance. DeepPruner \cite{duggal_deeppruner:_2019} reduces the dimension of cost volume step by narrowing down disparities range via a disparity estimation network. MobileStereoNet \cite{shamsafar2022mobilestereonet} and networks with separable convolutions  \cite{rahim2021separable} have focused on reducing the computational requirements of the stereo networks by exploiting the \mnet V1 and V2 modules of \cite{sandler2018mobilenetv2} . On the other hand, some of the recent methods have also focused solely on improving the overall performance of the stereo networks. \ganet \cite{zhang_ga-net:_2019} incorporates the successful traditional method \cite{hirschmuller_stereo_2008} in the form of a layer in an end-to-end network that improvs the performance but at the cost of high inference time. \acvnet \cite{xu2022acvnet} builts on \gwcnet and  improves its performance by deploying an attention-based cost volume. CFNet \cite{shen2021cfnet} uses cascade and fusion based cost volume to improve the performance. However majority of these methods have been targeting either performance or speed improvements independently. Thus there is a need of a method that can collectively merge the findings of these methods to build an efficient performant network.

%% file: methodology.tex
\input{arhc_figure}
We build our network based on the recent advancements and contributions in real-time image classification \cite{sandler2018mobilenetv2},\ object detection \cite{sandler2018mobilenetv2, wang2018pelee}, segmentation \cite{yu2021bisenet},  and performant stereo networks \cite{guo_group-wise_2019,xu2022acvnet} . Specifically, our architecture (\cf \figref{architecture}) consists of a shared backbone made of two branches -- \cf \secref{backbone}, an attention based cost volume and an hourglass based encoder-decoder network to do cost aggregation and disparity regression -- \cf \secref{cv-ed}. 

\subsection{Backbone Network}
\label{sec:backbone}
Our backbone network has been inspired by \mnet \cite{sandler2018mobilenetv2}, PeleeNet~\cite{wang2018pelee}, and \bisenet \cite{yu2021bisenet},  and aims at capturing both low-level features and semantic features present in the input image explicitly while being lightweight. More precisely, the backbone network consists of two branches called Shallow and Deep branches. 

\textbf{\SB:} This branch is responsible for capturing low-level image details. It consists of three levels and contains less number of layers compared to \db.  The exact design for \sb is present in \tabref{shallow}. Levels $ 1 $, $ 2 $ and $ 3 $  are build by downscaling the features by factor of $ 1/2 $, $ 1/4 $ and $ 1/8 $, respectively, compared to the original input height (H) and width (W). As this branch is responsible for capturing only fine details thus does not need larger receptive fields, therefore it is composed of only $ 8 $ convolutional layers.

\begin{table}[t]
	\centering
	\begin{adjustbox}{width=\linewidth}
		\def\arraystretch{1.5}
		\begin{tabular}{c|c|c|c|c|c}
			\hline
			\textbf{Level} &\textbf{Input Size} & \textbf{Operation} & \textbf{\makecell{Down Sample}} & \textbf{\makecell{Output Features}} & \textbf{Output Size}  \\ \hline
			1   & $ 3 \times H \times W $       & 3x3 Conv           & yes                  & 64                    & $ 64 \times H/2 \times W/2 $ \\ 
			1   & $ 64 \times H/2 \times W/2 $      & 3x3 Conv           & no                   & 64                    & $ 64 \times H/2 \times W/2 $ \\ \hline 
			2   & $ 64 \times H/2 \times W/2 $      & 3x3 Conv           & yes                  & 64                    & $ 64 \times H/4 \times W/4 $ \\ 
			2   & $ 64 \times H/4 \times W/4 $       & 3x3 Conv           & no                   & 64                    & $ 64 \times H/4 \times W/4 $  \\ 
			2   & $ 64 \times H/4 \times W/4 $       & 3x3 Conv           & no                   & 64                    & $ 64 \times H/4 \times W/4 $  \\ \hline
			3   & $ 64 \times H/4 \times W/4 $       & 3x3 Conv           & yes                  & 128                   & $ 128 \times H/8 \times W/8 $   \\ 
			3   & $ 128 \times H/8 \times W/8 $       & 3x3 Conv           & no                   & 128                   & $ 128 \times H/8 \times W/8 $    \\ 
			3   & $ 128 \times H/8 \times W/8 $       & 3x3 Conv           & no                   & 128                   & $ 128 \times H/8 \times W/8 $   \\ \hline
		\end{tabular}
	\end{adjustbox}
	\caption{Architecture of the \sb. H and W represent the height and width of the original input. Level correlates with dimensions of the feature maps \wrt original input size \eg for Level= 2, height of feature maps is H/($2^ {Level})$.}
	\label{table:shallow}
	\vspace*{-5pt}
\end{table}

\textbf{\DB:} This branch is responsible for capturing more higher-level information from the input images. To achieve this, the \db uses far more layers compared to the \sb and thus has higher a receptive field. With more layers, the backbone gets heavier. This is avoided by using fewer feature maps per layer. Moreover inside each layer, the \mnet \cite{sandler2018mobilenetv2} module is used to reduce the number of computations and keep the \db light-weight. More precisely, separable convolutions (\ie  replacing one $ 3 \times 3 $ convolution with a series of depth-wise separable and point-wise convolutions) are used to make the backbone deeper and increase the receptive field while limiting the amount of computations. The exact design of this branch can be found in \tabref{deeper}. Overall, this branch is built by adding a series  of Gather-and-Expansion (GE) \cite{yu2021bisenet} layers on top of a stem operation \cite{wang2018pelee}. The stem operation reduces the input size by a factor of $ 4 $,  whereas the series of GE further reduces the size of feature maps by a factor of $ 32 $. Finally,  a global average pooling layer at the end is used to capture the overall global context of the input for better semantic learning.  The final feature maps from this branch are at level $ 5 $ \ie $ H/32 \times W/32 $.\\
\begin{table}[t]
	\centering
	\def\arraystretch{1.5}
	\begin{adjustbox}{width=\linewidth}
		\begin{tabular}{c|c|c|c|c|c}
			\hline
			\textbf{Level} & \textbf{Input Size} & \textbf{Operation} & \textbf{Down Sample} & \textbf{\makecell{Output Features}} & \textbf{Output Size} \\ \hline
			1 \& 2 & $ 3 \times H \times W $      & Stem               & yes                  & 64                    & $ 16 \times H/4 \times W/4 $  \\ \hline
			3 & $ 16 \times H/4 \times W/4 $        & GELayer2           & yes                  & 32                    & $ 32 \times H/8 \times W/8 $ \\ 
			3 & $ 32 \times H/8 \times W/8 $        & GELayer1           & no                   & 32                    & $ 32 \times H/8 \times W/8 $  \\ \hline
			4 & $ 32 \times H/8 \times W/8 $        & GELayer2           & yes                  & 64                    & $ 64 \times H/16 \times W/16 $  \\ 
			4 & $ 64 \times H/16 \times W/16 $        & GELayer1           & no                   & 64                    & $ 64 \times H/16 \times W/16 $ \\ \hline
			5 & $ 64 \times H/16 \times W/16 $        & GELayer2           & yes                  & 128                   & $ 128 \times H/32 \times W/32 $ \\ 
			5 & $ 128 \times H/32 \times W/32 $        & GELayer1           & no                   & 128                   & $ 128 \times H/32 \times W/32 $ \\ 
			5 & $ 128 \times H/32 \times W/32 $        & GELayer1           & no                   & 128                   & $ 128 \times H/32 \times W/32 $ \\ 
			5 & $ 128 \times H/32 \times W/32 $        & GELayer1           & no                   & 128                   & $ 128 \times H/32 \times W/32 $ \\ 
			5 & $ 128 \times H/32 \times W/32 $        & Global Avg Pooling & no                   & 128                   & $ 128 \times H/32 \times W/32 $ \\ 
			5 & $ 128 \times H/32 \times W/32 $        & 1 x 1 Conv         & no                   & 128                   & $ 128 \times H/32 \times W/32 $ \\ 
			5 & $ 128 \times H/32 \times W/32 $        & 3 x 3 Conv         & no                   & 128                   & $ 128 \times H/32 \times W/32 $ \\ \hline
		\end{tabular}
	\end{adjustbox}
	\caption{Architecture of the \db.}
	\label{table:deeper}
	\vspace*{-5pt}
\end{table}
\textbf{Aggregation layer:} Aggregation layer is responsible for  fusing the information from both the branches in such a way that both low- and high-level information extracted is utilized maximally. A straightforward method like summation or a concatenation of the obtained feature maps cannot be used since the feature maps from both branches are on different scales. Therefore, like \cite{yu2021bisenet}, in aggregation layer features of one branch are used as attention weights to select the important features of the other branch. Selection of the features from the \sb is done in three steps. First a $ 3 \times 3 $ convolution and an upsampling operation on the output of the \db is applied to increase the spatial resolution of feature maps. Next these upsampled features maps are converted to attention weights by applying sigmoid. Finally these weights are element-wise multiplied by \sb features to select relevant features. Similarly for selecting \db features, feature maps from the \sb are first converted to attention weights after downsampling them to the same resolution of the \db. Next they are element-wise multiplied with \db output feature maps to select the important semantic information. Finally, selected shallow and \db features are added element-wise together after downsampling \sb features. The output resolution of aggregation layer is at level 5, \ie input resolution is reduced by factor of $2^5$, this helps to keep the later computations fast.
\subsection{Cost Volume, Cost Aggregation and Disparity Regression}
\label{sec:cv-ed}
\textbf{Cost Volume:} We construct an attention-based cost volume in the network, following the recent advancements \cite{xu2022acvnet}. Precisely, first a cost volume called `pre-attention volume' is constructed by concatenating left and right image features extracted via the backbone network. In parallel, another cost volume is constructed by taking group-wise correlation between left and right features.  This correlation volume is then split into three sub-groups and a series of convolution and merging operations are then applied to obtain attention weights. Finally the computed `attention weights'  are applied on `pre-attention volume' to filter the important disparities and give an `attention volume' as output for further processing -- \cf \figref{architecture}.\\

\textbf{Cost Aggregation:} Cost aggregation network consists of four \td convolution layers and two hourglass networks\cite{guo_group-wise_2019} stacked after one another. Each of the hourglass network has an encoder-decoder architecture with four \td convolutions (layers that reduce the input size by a factor of $ 2 $) and two \td transpose convolutions (to up-sample the feature maps).\\

\textbf{Disparity Regression:} Finally outputs from the three different points in the cost-aggregation network are used to regress disparities. These output points are: (i) the output of \td convolutions (Out$ 0 $); (ii)  the output of first hourglass network (Out$ 1 $); and the output (Out$ 2 $) of the second hourglass network -- \cf\figref{architecture}. Each of these outputs is passed through a pair of \td convolutions (to obtain a volume with a single feature map \ie $ 1 \times D/4 \times W/8 \times W/8 $)  and an upsampling layer (to achieve $ 1 \times D \times H \times W $ output). Next softmax in the disparity dimension is applied to convert these outputs to probabilities. Finally, \textit{argsoftmax} in disparity dimension is applied to get disparity values. Note that during training all the three different disparities values computed are used in the loss function calculation whereas in inference only the Out$2$ is used to regress the final disparities -- \cf \figref{architecture}.

\subsection{LogLoss}
\label{sec:loss}
Stereo methods typically use the L2, L1 or its variants like SmoothL1 as loss function. Generally speaking, \ie
\begin{equation}
	loss = \frac{\sum_{i=1}^{H} \sum_{j=1}^{W} E(p_{ij},q_{ij})}{H \times W}
\end{equation}
where \textit{p} is the predicted disparity, \textit{q} is the ground-truth disparity, \textit{ij} represents a pixel and $E$ defines how the error is computed between the predicted and ground-truth disparities, \eg in case of L1 loss $E(x,y)=|x-y|.$ 

If we consider the partial derivatives of the loss function with respect to networks parameters $\theta$, then we see that for L1 it is 
\begin{equation}
	\frac{\partial loss}{\partial \theta}= \frac{\partial E}{\partial x} \frac{\partial x}{\partial \theta}, \mbox{where} 
\end{equation}
\begin{equation*}
	\frac{\partial E}{\partial x}=
	\begin{cases}
		1 & \text{if $x-y > 0$}\\
		-1 & \text{else}\
	\end{cases}       \\
\end{equation*} and for  L2 it is 
\begin{equation*}
	\frac{\partial E}{\partial x}= 2(x-y) 
\end{equation*}
Thus these error functions either have constant influence on the learning (in case of L1) or proportional to error (in case of L2) without considering whether the error computed involves outliers or not. 
Alternatively, for small error differences either there are very small adjustments or constant adjustments to learning.
Here our goal is to have a loss function that gives more importance to the small-errors, thus following the works in mono-disparity estimation \cite{Hu2018,watson2019,Jae-Han2018single} we also propose to use $\log$ ($\frac{\partial E}{\partial x}=\frac{1}{x}$, where $\frac{\partial E}{\partial x} \rightarrow 0, x \rightarrow \inf$) as a squeezing function that penalizes small disparity errors more in proportion to the large disparity errors. Precisely we define, 
\begin{equation}
	E=log(|p_{ij}-q_{ij}|+\epsilon)
\end{equation}
%
We set $\epsilon =1$ to keep the loss positive. 
In our experiments, we found that not only does this loss function lead to better performance but is also more stable during the training.

%

%% file: arhc_figure.tex
\begin{figure*}[!ht]
	\centering
	\includegraphics[width=\linewidth]{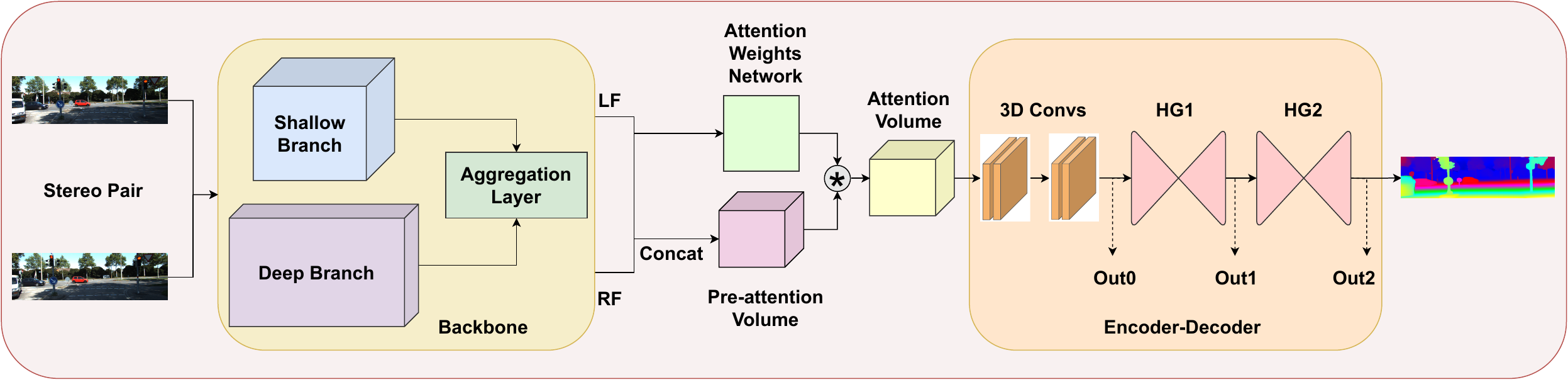}
	\caption{Proposed Architecture. Here LF and RF represent the left and right image features extracted via the backbone network (`Concat` means concatenation of LF and RF). HG1 and HG2 represent the first and second hour-glass of the cost aggregation module. Out0, Out1 and Out2 are the outputs from cost aggregation and are used to regress the disparities.}
	\label{fig:architecture}
		\vspace*{-10pt}
\end{figure*}

%% file: experiments-results.tex
In this section, we first introduce the datasets used for the training and evaluation of our experiments. We then present the implementation details, followed by different design choices and results.
\subsection{Datasets}
We train and evaluate our models on two well-known stereo benchmarks, namely \sf \cite{mayer_large_2016} and \kittif ~\cite{menze_object_2015}.

\textbf{\kitti} datasets introduced as a benchmark are real world datasets of driving scenes \cite{geiger2012we, menze_object_2015}. \kittit contains 389 image pairs ($ 194 $ training and $ 195 $ test image pairs) while \kittif contains  400 ($200$ training and $200$ test image pairs) image pairs of resolution $376 \times 1240$ pixels with sparse ground-truth disparities for learning and evaluating stereo algorithms. 


\textbf{\sf} is a large scale synthetic dataset introduced by Mayer \etal \cite{mayer_large_2016}. It contains $35,454$ training images and $4,370$ test images of size $540 \times 960$ with dense ground truths. 

\textbf{Evaluation metrics:}  We report evaluation metrics including \dpix, \tpix, \opix, D1 and End-Point-Errors (EPE)~\cite{mayer_large_2016,menze_object_2015}.  EPE (reported in \textit{px} units) is the average disparity error whereas \dpix represents the percentage of pixels whose error is greater than $ 3 $ pixels. Similarly, \tpix and \opix, are percentage of pixels with errors greater than $ 2 $ and $ 1 $ pixels, respectively. D1 is the percentage of pixels where the \dpix error is greater than  $ 0.05$ $ \times $ ground-truth. We also report the number of parameters in millions (M) and number of \macs (Multiply-ACcumulate operations) in Giga-MACs (GMACss) for comparison. Note that $ 1  \mbox{MAC} =  1 $ multiplication + $ 1 $ addition operation.
\subsection{Implementation details}
We chose \gwcnet \cite{guo_group-wise_2019} as our baseline method.  For training, we use the Adam optimizer ($ \beta_1=0.9  $ and $ \beta_2=0.999 $). All the training images are cropped to the size of $256 \times 512$ with the maximum disparity set to $192$. Furthermore, all the images are standardized using the mean and standard deviation of the ImageNet dataset. Our final model is trained with a batch size of $ 8 $ on a set of $ 4 $ GPUs. For the \sf dataset, we train our model for 900K iterations with an initial learning rate of \textit{1E-3} which is reduced to half at fixed iterations steps, \ie, at  600K, 700K, and 800K.
For the \kittif dataset, we fine-tuned our pre-trained sceneflow model on KITTI datasets for 11K iterations with batch size of $ 16 $. The initial learning rate was set to \textit{1E-3} which was later decreased to \textit{1E-4} after 7K iterations. 
\subsection{Evaluation of design choices}
\textbf{Effect of light-weight backbones:} During the evolution of backbone we trained and experimented with different other light-weight backbones like  ResNet18, ResNet34 \cite{he2016deep}, MobileNet \cite{sandler2018mobilenetv2}, \bisenet, and other efficient network building blocks.  More specifically, we experimented by replacing the backbone in our baseline with ResNet18 \cite{he2016deep}, ResNet34 \cite{he2016deep}, MobileNet \cite{sandler2018mobilenetv2} and \bisenet \cite{yu2021bisenet} and its variants. Although ResNet18, ResNet34, MobileNet backbone require significantly less number of operations than \bisenet (\ie $ 5.1\times, 2.9\times$ and $13.6\times$ less operations respectively) however their performances were also far worse, \eg ResNet34 had an EPE of 2 as compared to EPE of 1.3 of our \bisenet influenced model.  Therefore we chose \bisenet building blocks as our backbone, as this seems a reasonable choice \wrt to performance and speed. This is somewhat understandable as well since \bisenet is optimized specially for image segmentation task which is itself a pixel-level task and thus have inherent capabilities to capture the semantic, textual and structural information at pixel-level.  For \bisenet we also experimented with different sizes of \sb and \db, with different number of filters, number of layers and finally we settled with architecture reported in the Tables \ref{table:shallow} and \ref{table:deeper}.
\textbf{Effect of loss function:}
In our initial experiments we did a comparison between L2, L1 and SmoothL1 loss but SmoothL1 turned out to be  consistently better. However the difference in performance gap to other competing methods were still large especially \opix error -- \cf \tabref{loss-func}. We observed that despite of using SmoothL1 loss the magnitude of \tpix and \opix errors are high. As soon as we replaced the loss function with \lossname, we observed a noticeable performance improvement -- \cf  \tabref{loss-func}.  Here we can observe that  \lossname not only improves EPE but also improves all others errors significantly, especially \opix error. In addition we also observed that \lossname leads to faster convergence and stable training of models.\\
\begin{table}[]
	\centering
	\begin{adjustbox}{width=0.9\columnwidth}
		\npdecimalsign{.}
		\nprounddigits{2}
		\begin{tabular}{l|n{2}{3}|n{2}{2}|n{2}{2}|n{2}{2}|n{2}{2}}
			\hline
			\textbf{Loss function} & {EPE($px$)} & {D1($\%$)} & {\dpix($\%$)} & {\tpix($\%$)} & {\opix($\%$)}\\ \hline
			L2 & 1.3048 & 6.524 & 7.885 & 12.033 & 24.276\\ 
			SmoothL1 & 0.8262 & 2.856 & 3.58 & 5.24 & 10.796 \\  
			LogL1Loss & 0.749 & 2.47 & 3.015 & 3.888 & 6.326\\
			\hline
		\end{tabular}
	\end{adjustbox}
	\caption{Performance evaluation of different loss functions.}
	\label{table:loss-func}
	\vspace*{-5pt}
\end{table}
\textbf{Effects of adding attention in the cost volume:}
\tabref{cost-volume} compare the performance of the model trained with a simple group-wise correlation volume and a model where additional attention weights are learned and used to filter the important disparities. We can observe that the model with attention weights based  cost volume improves the performance for all the metrics. Overall, adding attention leads to slight increase in MACs but the performance gains far outweight the increase in MACs here. \\
\begin{table}[]
	\centering
	\begin{adjustbox}{width=0.9\columnwidth}
		\npdecimalsign{.}
		\nprounddigits{2}
		\begin{tabular}{l|n{2}{3}|n{2}{2}|n{2}{2}|n{2}{2}|n{2}{2}}
			\hline
			\textbf{Cost Volume} & {EPE($px$)} & {D1($\%$)} & {\dpix($\%$)} & {\tpix($\%$)} & {\opix($\%$)} \\ \hline
			w/o Attention & 0.8602 & 2.944 & 3.546 & 4.569 & 7.405\\ 
			w Attention & 0.749 & 2.47 & 3.015 & 3.888 & 6.326\\
			\hline
		\end{tabular}
	\end{adjustbox}
	\caption{Effect of using learned attention weights for refining cost volume.}
	\label{table:cost-volume}
	\vspace*{-5pt}
\end{table}
\textbf{Effect of Separable Convolutions in Encoder-Decoder:}
To further reduce the computational requirements of our network we considered replacing costly \tdcs in encoder-decoder with light-weight depth-wise separable convolutions as suggested in \cite{rahim2021separable}. \tabref{convs} shows that although \td separable convolutions reduces the number of parameters by $ 2\times $ but we also witness a noticeable performance drop in all metrics. Furthermore, since \td separable convolutions do not have optimized kernel operations therefore, for our network we decide  not to use \tdcs in the encoder-decoder part.
\begin{table}[]
	\centering
	\begin{adjustbox}{width=\columnwidth}
		\npdecimalsign{.}
		\nprounddigits{2}
		\begin{tabular}{l|c|c|c|c|c|c|c}
			\hline
			\makecell{\textbf{Encoder-} \\ \textbf{decoder}} & \makecell{EPE\\($px$)} & \makecell{D1\\($\%$)} & \makecell{\dpix\\($\%$)} & \makecell{\tpix\\($\%$)} & \makecell{\opix\\($\%$)} & \makecell{MACs\\($G$)} & \makecell{Params\\($M$)} \\ \hline
			\makecell{3D Separable Conv} & 0.92 & 3.04 & 3.65 & 4.73 & 7.85 & 4.95 & 36.51\\ 
			3D Conv & 0.75 & 2.47 & 3.02 & 3.89 & 6.33 & 6.6 & 66.03\\
			\hline
		\end{tabular}
	\end{adjustbox}
	\caption{3D Convolutions vs Depth-wise Separable 3D Convolutions in encoder-decoder.}
	\label{table:convs}
		\vspace*{-10pt}
\end{table}

\textbf{Effect of the optimizer:}
We evaluated the impact of different optimizers on our network performance. Specifically, we trained our models with Adam \cite{kingma2014adam}, AdamW \cite{loshchilov2017decoupled} and SGD optimizer. For our experiment, Adam and AdamW showed similar results EPE 0.75 and 0.76 respectively while SGD gave little inferior performance 0.80 EPE. Therefore all the later networks were trained using Adam optimizer.\\

Based on conclusions drawn from the experiments on different design choices, we train our leaner backbone network with attention-based cost volume using {\lossname}. This network is optimized using Adam optimizer with a starting learning rate of \textit{1E-3}.
\subsection{Results}
This section presents our quantitative and qualitative results on test and validation sets of \kittif and \sf.

\input{results/sceneflow_quanti}
\textbf{\sf:} 
\tabref{sceneflow} shows our quantitative results on the \sf test set, compared to existing stereo methods. We can observe that our network has superior EPE and D1 error than \gwcnet and \leastereo while having $ 4\times $ and 2.5 $\times$ less operations, respectively. In addition, our network also contains $ 3.6\times $ fewer operations when compared with ACVNet \cite{xu2022acvnet} -- for timing comparison \cf \secref{inference-time}.

\figref{SceneFlow} shows qualitative results of our method in comparison with other existing methods. From the disparity errors we can see that our method gives reliable disparity estimates that appear better than \gwcnet while being comparable with \acvnet ones.

\input{results/sceneflow_quality}
\textbf{\kittif:}
For the \kitti dataset, we report both validation and test set results. All of the methods reported in \tabref{kitti2015} are trained by us, with the same train/validation split for a fair comparison. We can observe from the reported results that the performance of our model is on par with other methods, while having significantly fewer computations; for example, GwcNet and GANet require $ 4\times $ and $ 11\times $ more operations than our model, leading to their increased inference time in comparison.
\input{results/kitti2015}
We also submitted the results of our model on the \kitti benchmark. \tabref{3DKITTI2015} shows the quantitative comparison with other benchmark methods. We report the D1 error for foreground \textit{(fg)}, background \textit{(bg)} and all pixels following the benchmark. 
Qualitative results on \kitti benchmark are shown in \figref{kitti-benchmark-quality}.
\input{results/3DKITTI2015}
\input{results/KittiBenchmark}
\subsection{Inference time comparison}\label{sec:inference-time}
\tabref{memory-time} shows the inference time comparison of our final optimized method
with 2D, 3D and other real time methods. In the right table all the results are reproduced from the original papers\footnote{As no public or open-source implementations for the majority of these methods are available.} and thus might not be uniformly comparable across different image-resolutions and different types of hardwares. In the left table we reproduce the results using the following experimentation protocol.
\input{results/memorytime}

In this protocol, we follow a two-step process to measure the inference timings. In the first step, a warm-up cycle is performed using a subset of images (20 images of size $512\times960$\footnote{rounded to near values for models that take input as a factor of $X$.}) to warmup the GPU. In the next step a larger subset of images (400 images of size $512\times960$) is used to record the mean  inference time with the standard deviation. This whole two-step process is repeated ``k" times and then the average of the ``k=3" runs is reported. 

We can observe that our optimized  network is  $ 17\times $,  $\sim 9\times $, $27\times$, $14\times $ faster than state of the art methods like \psm\cite{chang_pyramid_2018},  \acvnet \cite{xu2022acvnet}, \leastereo~\cite{cheng2020hierarchical} and \cfnet~\cite{shen2021cfnet} respectively. Moreover, our method has better run-time and EPE relative to all the compared 2D and real-time methods.  

%% file: results/sceneflow_quanti.tex
\begin{table}[tbp]
	\begin{adjustbox}{width=\columnwidth} 
			\centering		
			\begin{tabular}{l|cccc|cc}		
				\hline
				{Method} & \makecell{EPE\\($px$)} & \makecell{D1\\($\%$)} & \makecell{MACs\\($G$)} & \makecell{Params\\($M$)} & \makecell{More \\ MACs}\\
				\hline
				GCNet\cite{kendall_end--end_2017}      	   			& 1.84 & - & 718.01 & 3.18    & 10.9x\\
				PSMNet\cite{chang_pyramid_2018}		  			 & 0.88 & 3.48 & 256.66 & 5.22    & 3.9x \\
				GA-Net-deep\cite{zhang_ga-net:_2019}	 		   			& 0.84 & 3.29 & 670.25 & 6.58    & 10x \\
				GA-Net-11\cite{zhang_ga-net:_2019}			  		 & 0.93 & 3.49 & 383.42 & 4.48    & 5.8x \\
				GwcNet-g\cite{guo_group-wise_2019}			   		& 0.79 & 2.77 & 246.27 & 6.43    & 3.7x \\
				DeepPruner\cite{duggal_deeppruner:_2019} & 0.86  & - & 129.23 & 7.39& 2.0x \\ 
				CFNet\cite{shen2021cfnet} & 0.97  & 4.5 & 177.50 & 23.05& 2.7x \\ 
				LEAStereo\cite{cheng2020hierarchical} & 0.78  & - & 156.5 & {1.81} & 2.4x \\ 
 \hline
				\methodname (Ours) & 0.75 & 2.47 & {66.03} & 6.6  & - \\ \hline 
			\end{tabular}
			\vspace*{-0.5cm}
	\end{adjustbox}
	\caption{Comparison on SceneFlow test set.
Qualitative comparison of \methodname on the SceneFlow test set with competing state of the art methods \wrt error metrics, parameters and \macs. Note that	despite having fewer \macs and faster run-time our proposed method gives similar performance compared to other state of the art methods.
}
	\label{table:sceneflow}	
\end{table}

%% file: results/sceneflow_quality.tex
\begin{figure*}[t]
	\centering
	\begin{tabular}{cccc}
		\includegraphics[width=0.23\linewidth]{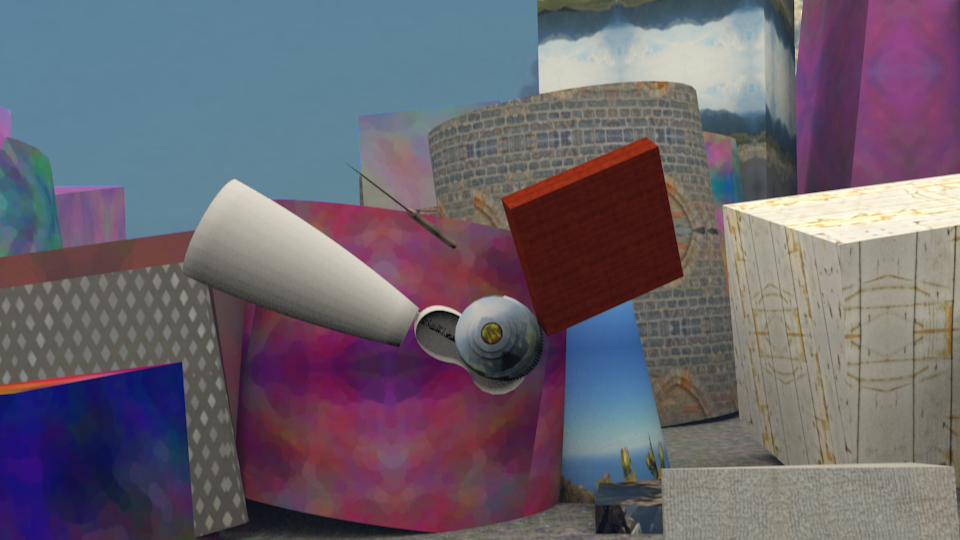} &	\includegraphics[width=0.24\linewidth]{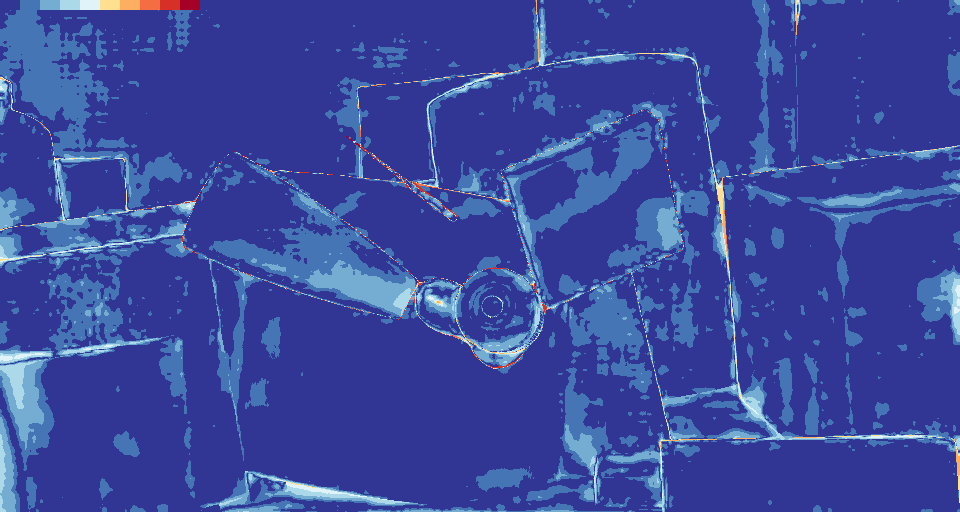} &	\includegraphics[width=0.24\linewidth]{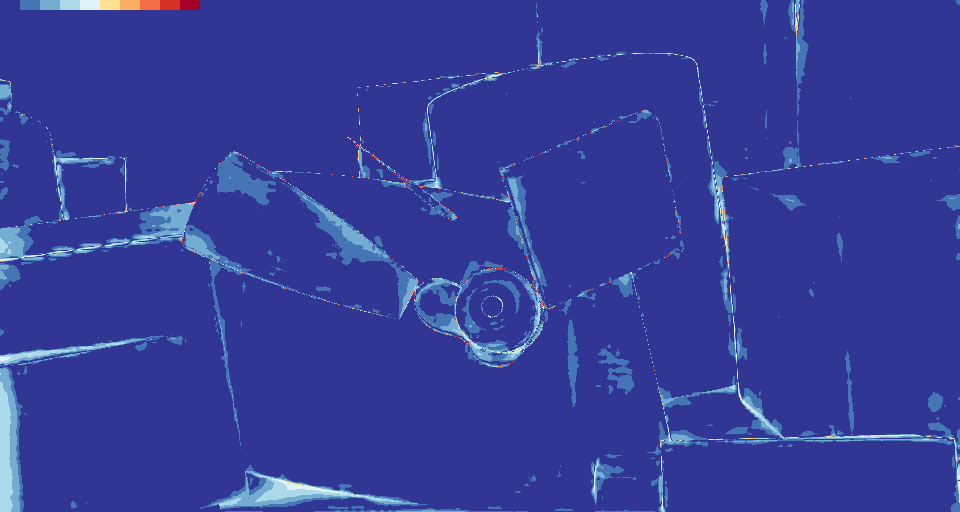} &	\includegraphics[width=0.24\linewidth]{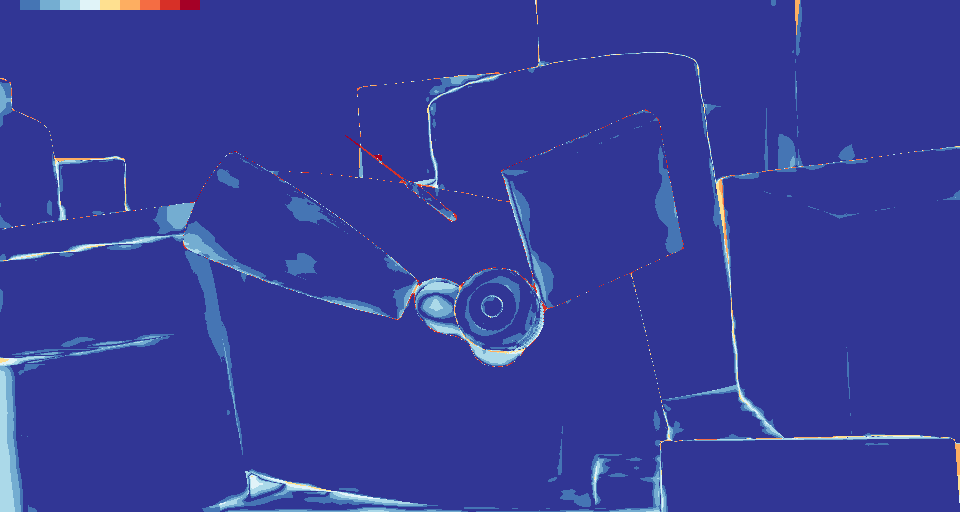} 
		\\
		\includegraphics[width=0.23\linewidth]{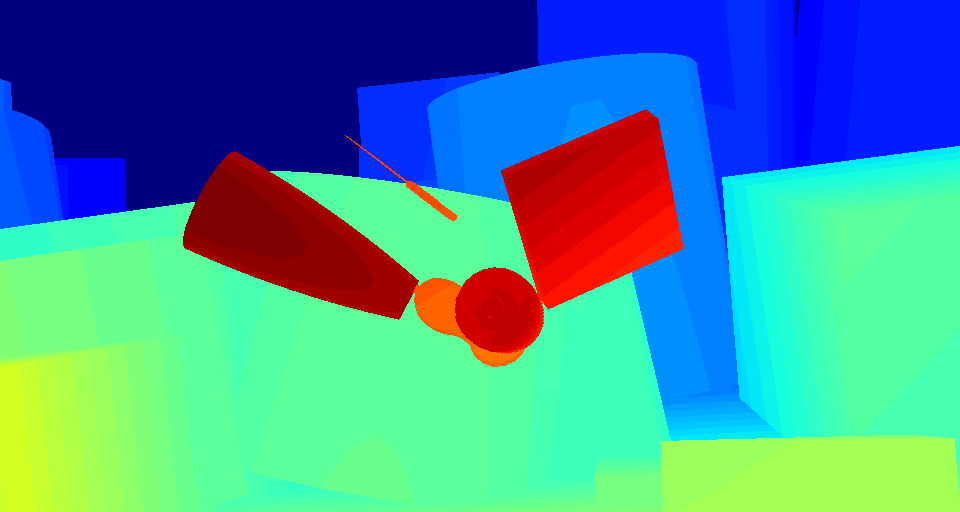} &	\includegraphics[width=0.24\linewidth]{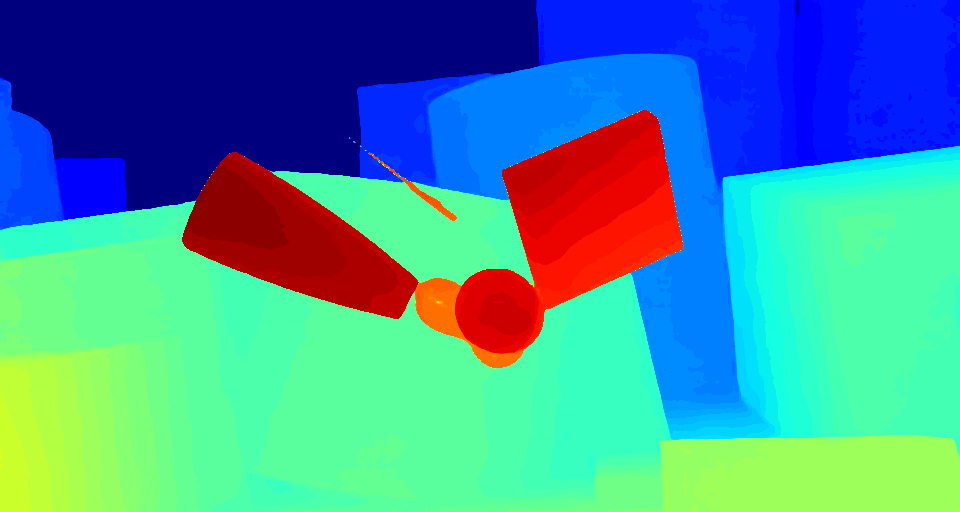} &	\includegraphics[width=0.24\linewidth]{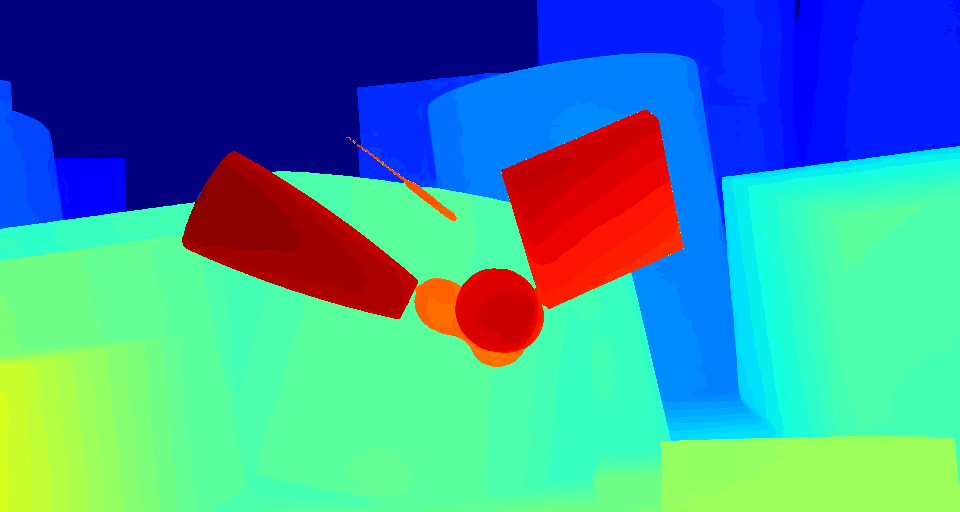} &	\includegraphics[width=0.24\linewidth]{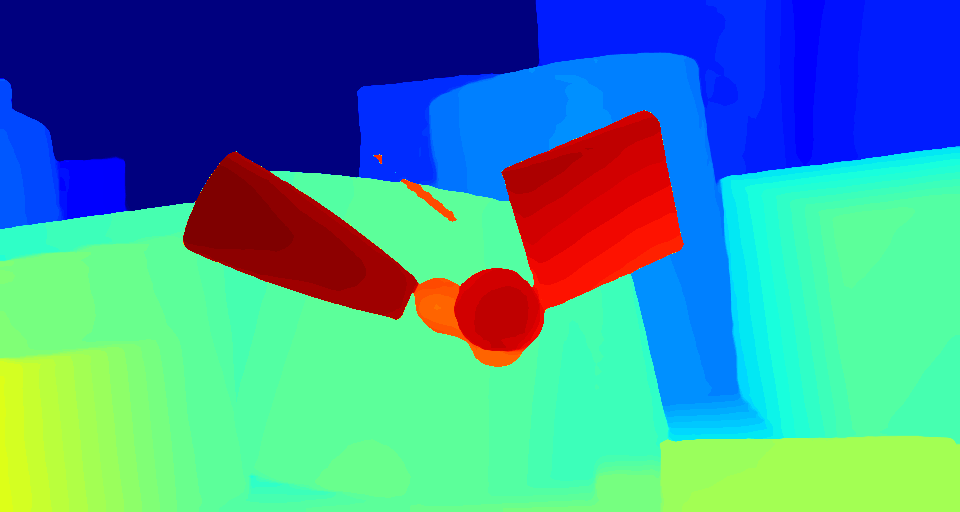}
		\\
		\includegraphics[width=0.23\linewidth]{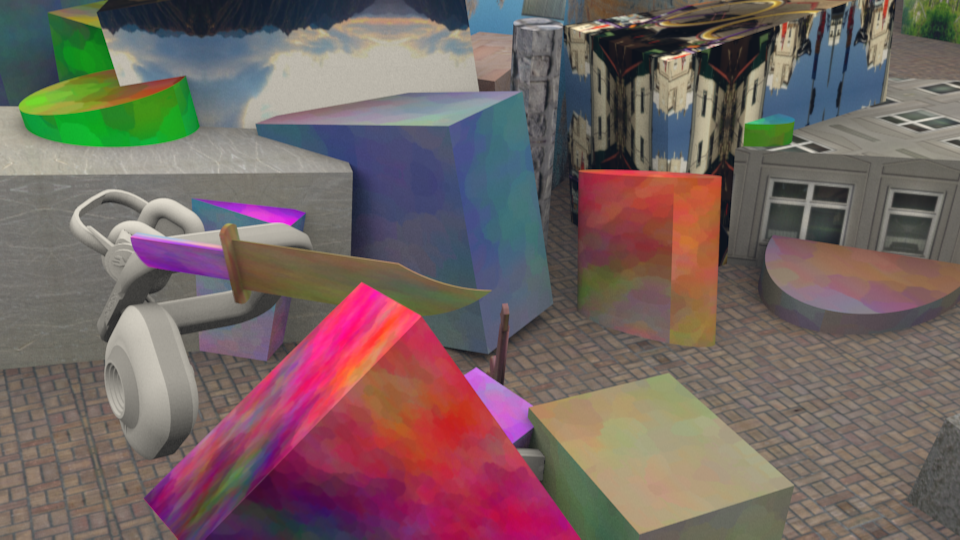} &	\includegraphics[width=0.24\linewidth]{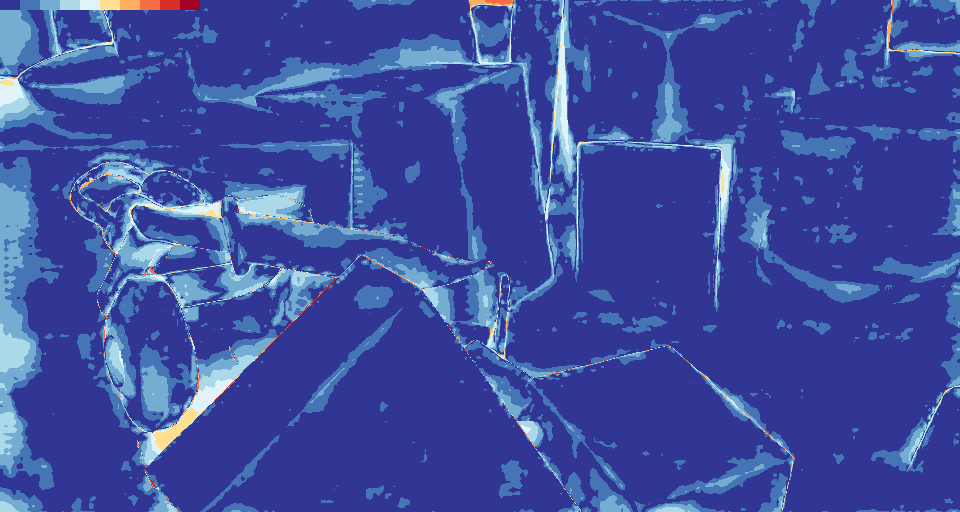} &	\includegraphics[width=0.24\linewidth]{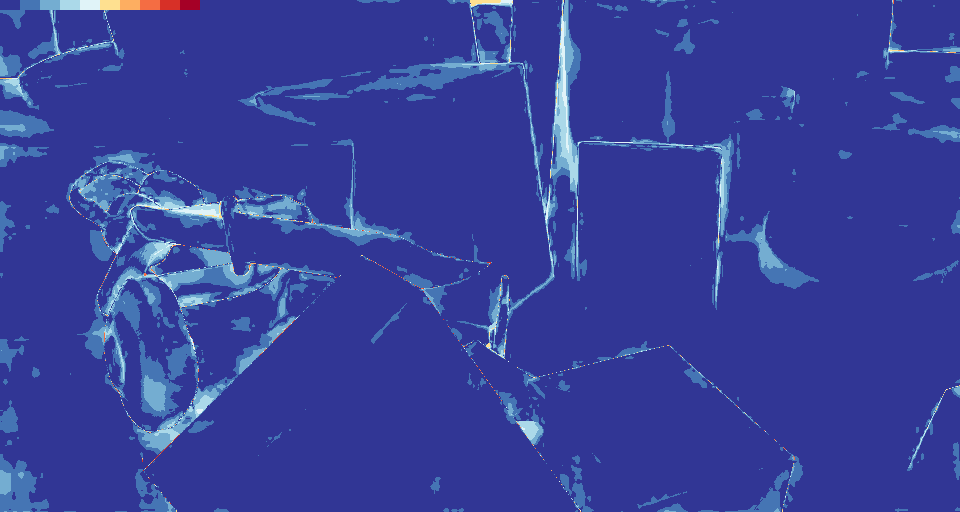} &	\includegraphics[width=0.24\linewidth]{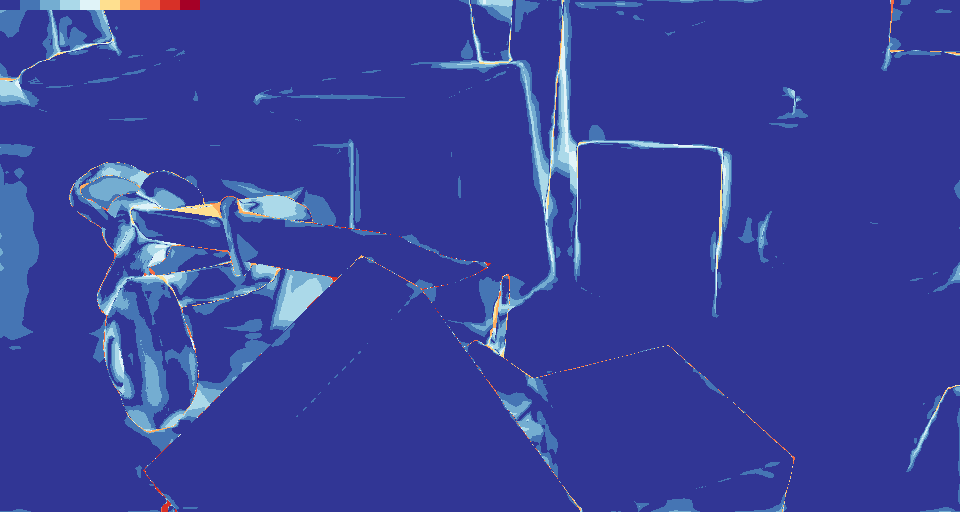} 
		\\
		\includegraphics[width=0.23\linewidth]{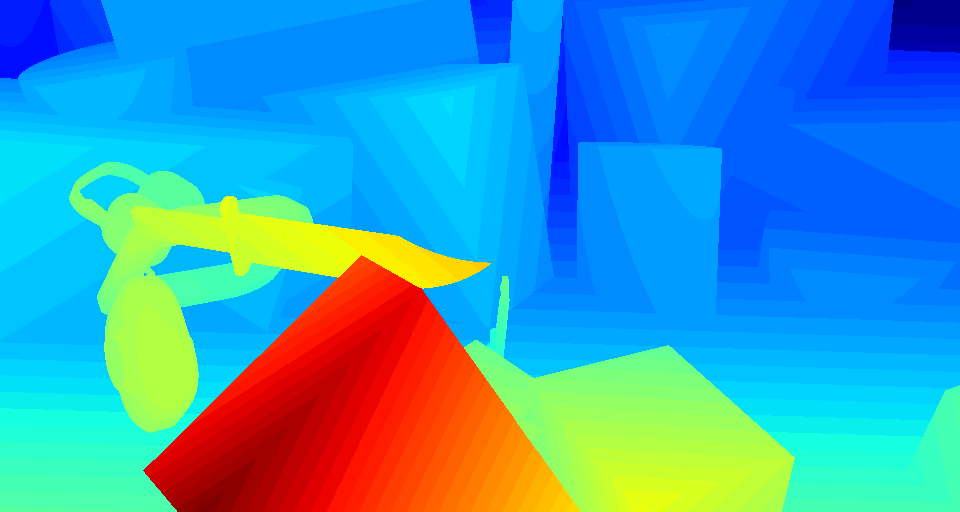} &	\includegraphics[width=0.24\linewidth]{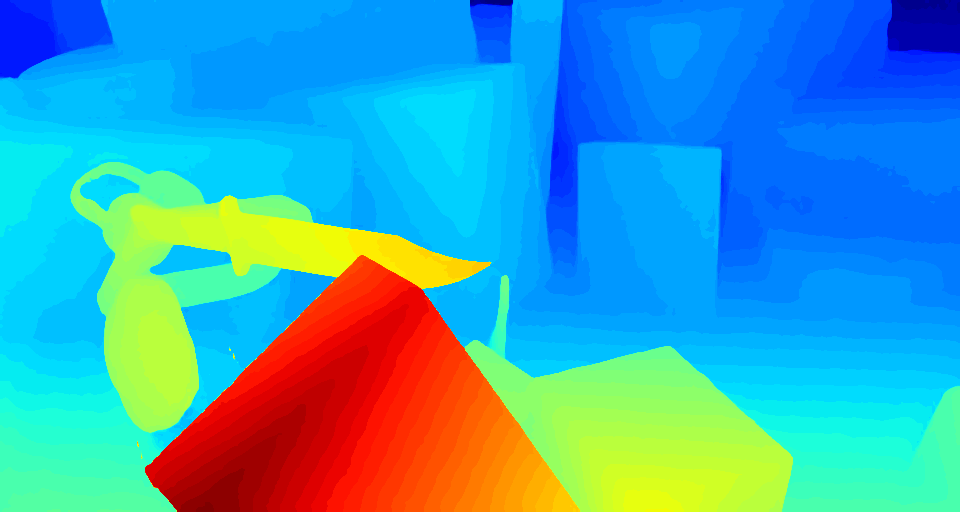} &	\includegraphics[width=0.24\linewidth]{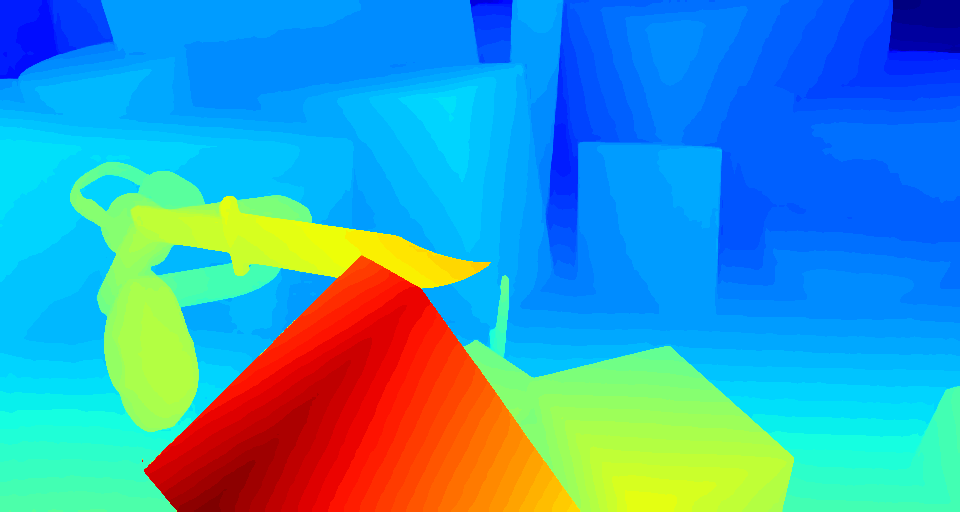} &	\includegraphics[width=0.24\linewidth]{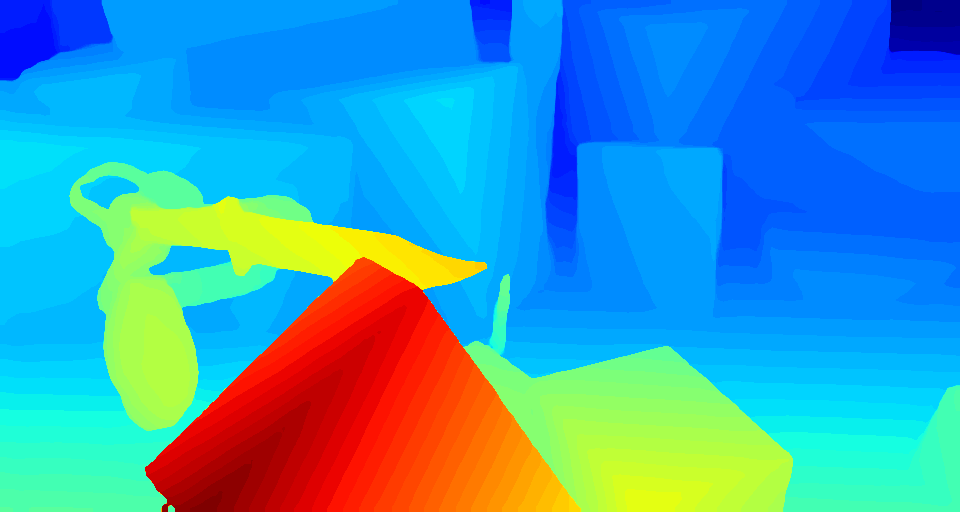}
		\\
		Left Image / GT & GwcNet-g\cite{guo_group-wise_2019} & ACVNet\cite{xu2022acvnet} & \methodname (Ours)  \\
	\end{tabular}
	\caption{Qualitative results on sample SceneFlow images.	Here the first and third rows represent the error maps w.r.t. ground truth. Darker red and blue colors represent higher and lower disparity errors, respectively.}
	\label{fig:SceneFlow}
	\vspace*{-15pt}
\end{figure*}

%% file: results/kitti2015.tex
\begin{table}[tbp]
	\begin{center}
		\footnotesize
		\begin{tabular}{@{\hskip2pt}l@{\hskip1pt}|@{\hskip2pt}c@{\hskip2pt}c@{\hskip2pt}c@{\hskip2pt}|@{\hskip2pt}c@{\hskip2pt}c@{\hskip2pt}}
			\hline				
			{Method} & {EPE($px$)} & {D1(\%)} & {\dpix(\%)} & {MACs($G$)} & {Params($M$)} \\				
			\hline				
			PSMNet\cite{chang_pyramid_2018} & 0.88 & 2.00 & 2.10 & 256.66 & 5.22 \\
			GA-Net-deep\cite{zhang_ga-net:_2019} & 0.63 & 1.61 & 1.67 & 670.25 & 6.58 \\
			GA-Net-11\cite{zhang_ga-net:_2019}  & 0.67 & 1.92 & 2.01 & 383.42 & \textbf{4.48} \\
			GwcNet-gc\cite{guo_group-wise_2019} & 0.63 & 1.55 & 1.60 & 260.49 & 6.82 \\
			GwcNet-g\cite{guo_group-wise_2019}  & \textbf{0.62} & \textbf{1.49} & \textbf{1.53} & 246.27 & 6.43 \\\hline
			\methodname  (Ours)  & \textbf{0.62} & 1.78 & 1.82 & \textbf{60.41}  & 6.60 \\			
			\hline				
		\end{tabular}
	\end{center}
	\vspace*{-0.2cm}
	\caption{Quantitative comparison of \methodname on the \kittif validation set with competing state of the art methods.}
	\label{table:kitti2015}
\end{table}

%% file: results/3DKITTI2015.tex
\begin{table}[tbp]
	\begin{center}
		\footnotesize
		\begin{tabular}{@{\hskip5pt}l@{\hskip5pt}|@{\hskip5pt}c@{\hskip5pt}c@{\hskip5pt}c@{\hskip5pt}|@{\hskip5pt}c@{\hskip5pt}c@{\hskip5pt}c@{\hskip5pt}}
			\hline
			\multirow{2}{*}{Methods} & \multicolumn{3}{c|}{All(\%)} & \multicolumn{3}{c}{Noc(\%)}  \\	 \cline{2-7} 
			& {D1$_{bg}$} & {D1$_{fg}$} & {D1$_{all}$} & {D1$_{bg}$} & {D1$_{fg}$} & {D1$_{all}$}  \\ \hline
			MC-CNN\cite{zbontar_stereo_2016} & 2.89 & 8.88 &3.89 & 2.48 & 7.64& 3.33\\
			Fast DS-CS\cite{yee2020fast} & 2.83 &	4.31	& 3.08 & 2.53&	3.74&	2.73 \\
			GCNet\cite{kendall_end--end_2017} & 2.21 & 6.16 & 2.87 & 2.02 & 5.58 & 2.61 \\
			PSMNet\cite{chang_pyramid_2018} & 1.86 & 4.62 & 2.32 & 1.71 & 4.31 & 2.14 \\
			AutoDispNet-CSS\cite{saikia_autodispnet:_2019} &1.94	& 3.37	&2.18 &1.80	& 2.98	&2.00\\
			DeepPruner\cite{duggal_deeppruner:_2019} & 1.87 & 3.56 & 2.15 & 1.71 & 3.18 & 1.95 \\ 
			GwcNet-g\cite{guo_group-wise_2019} & 1.74 & 3.93 & 2.11 & 1.61 & 3.49 & 1.92 \\
			CFNet\cite{shen2021cfnet} & {1.54} & 3.56 & {1.88} & {1.43} & 3.25 & {1.73} \\
			AANet\cite{xu_aanet_2020} & 1.99 & 5.39 & 2.55 & 1.80 & 4.93 & 2.32 \\
			\acvnet\cite{xu2022acvnet} & 1.37 & 3.07 & 1.65 & 1.26 & 2.84 & 1.52 \\\hline
			\methodname (Ours)  &  1.87 &	3.51 &	2.15 &  1.74 &	3.15 &	1.97  \\
			\hline
		\end{tabular}
	\end{center}
	\vspace*{-0.2cm}
	\caption{Qualitative comparison of \methodname on the \kittif benchmark with competing state of the art methods.}
	\label{table:3DKITTI2015}
			\vspace*{-0.5cm}
\end{table}

%% file: results/KittiBenchmark.tex
\begin{figure*}
	\footnotesize
	\begin{tabular}{ccc}
		\includegraphics[width=0.30\linewidth]{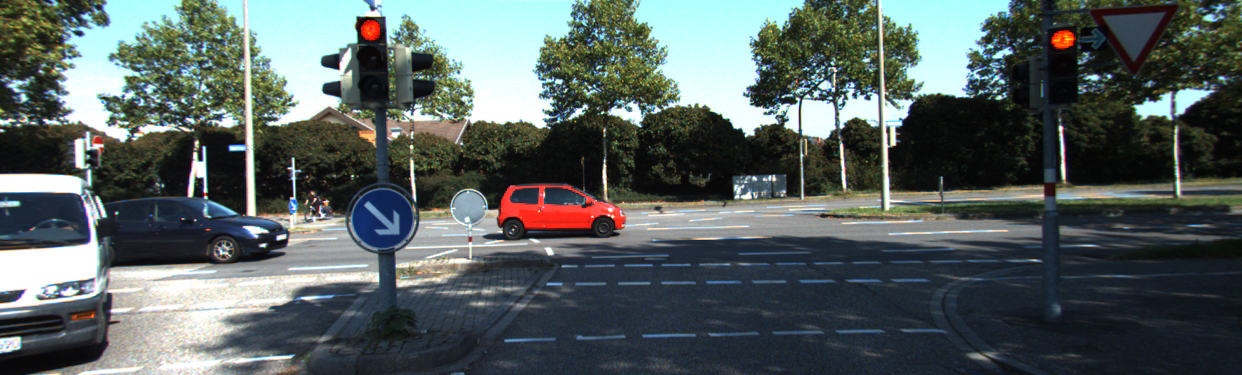} & \includegraphics[width=0.30\linewidth]{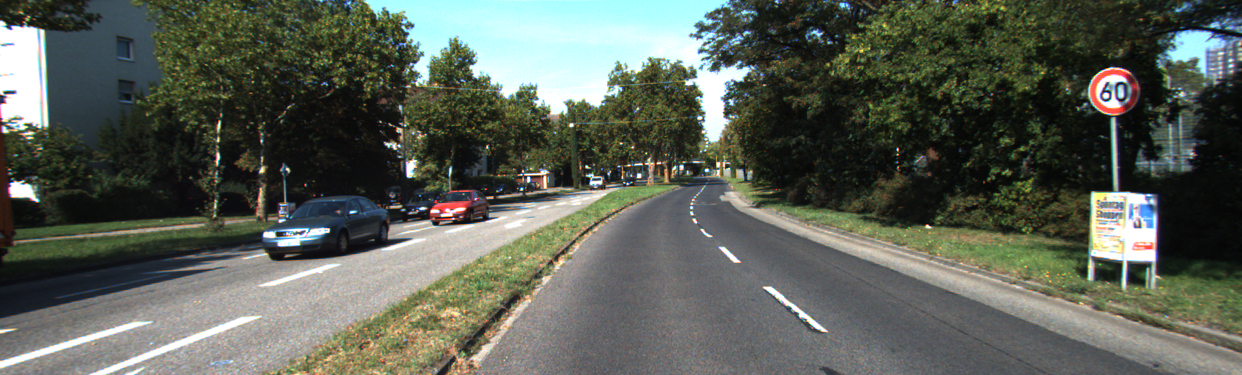} & \includegraphics[width=0.30\linewidth]{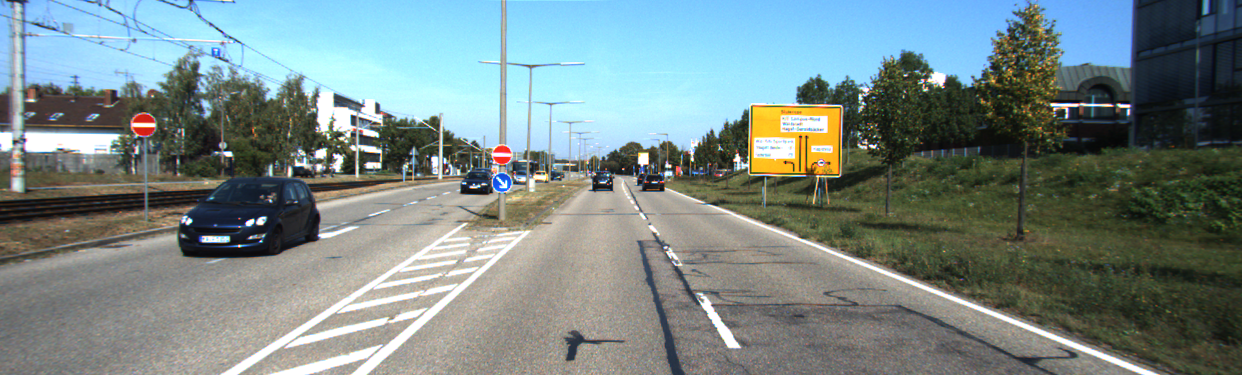} \\
		& Left Images & \\
		
				\includegraphics[width=0.30\linewidth]{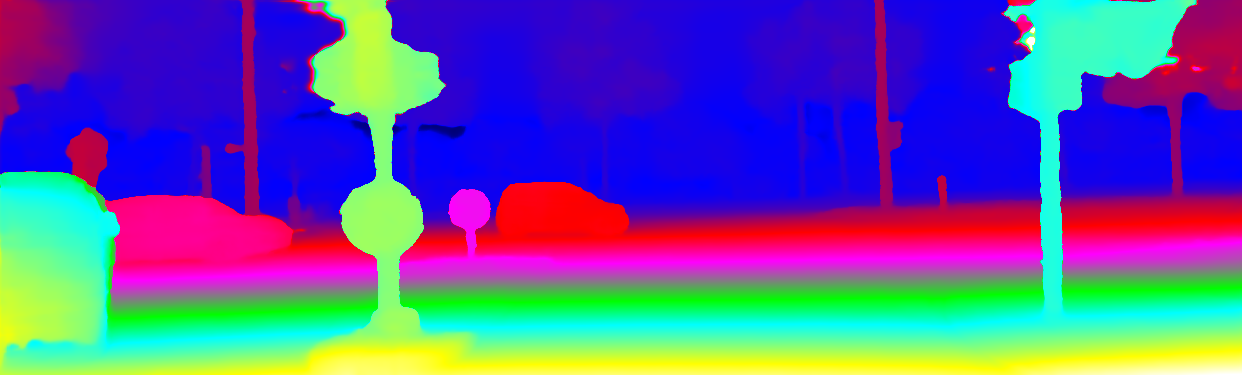} & \includegraphics[width=0.30\linewidth]{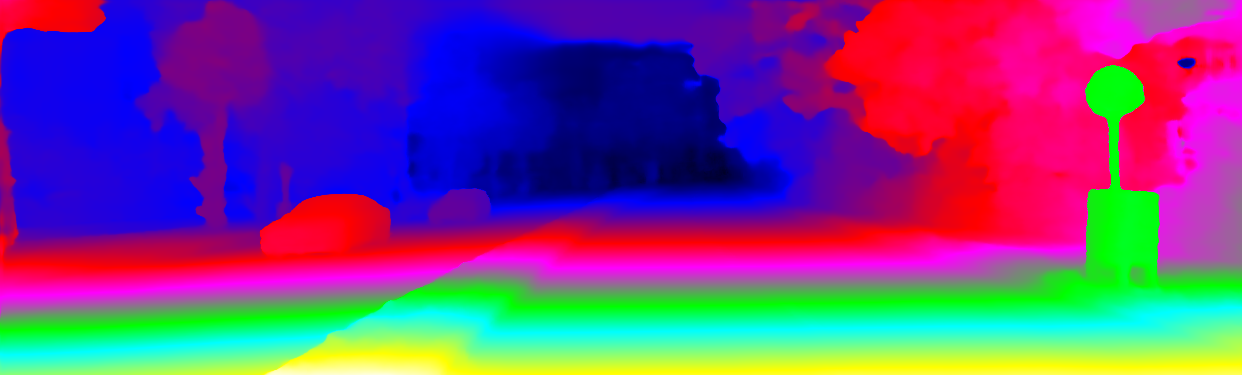} & \includegraphics[width=0.30\linewidth]{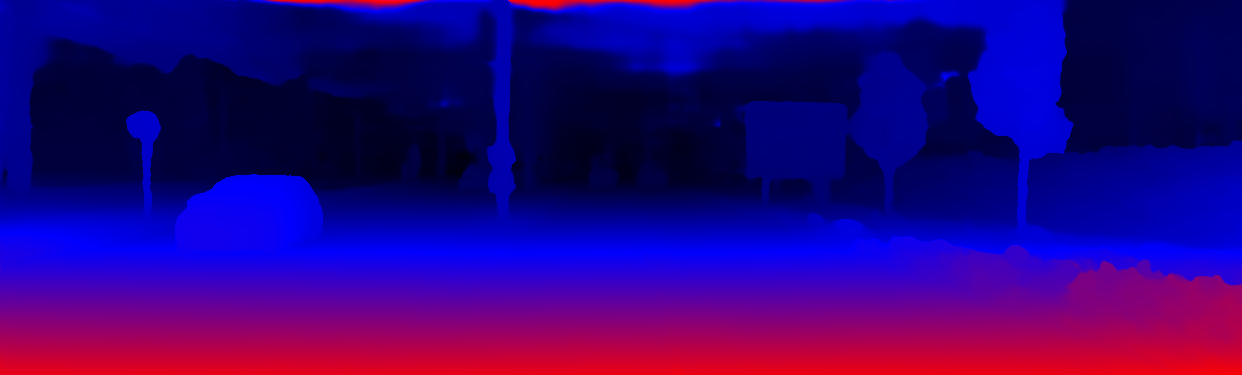} \\
				
				\includegraphics[width=0.30\linewidth]{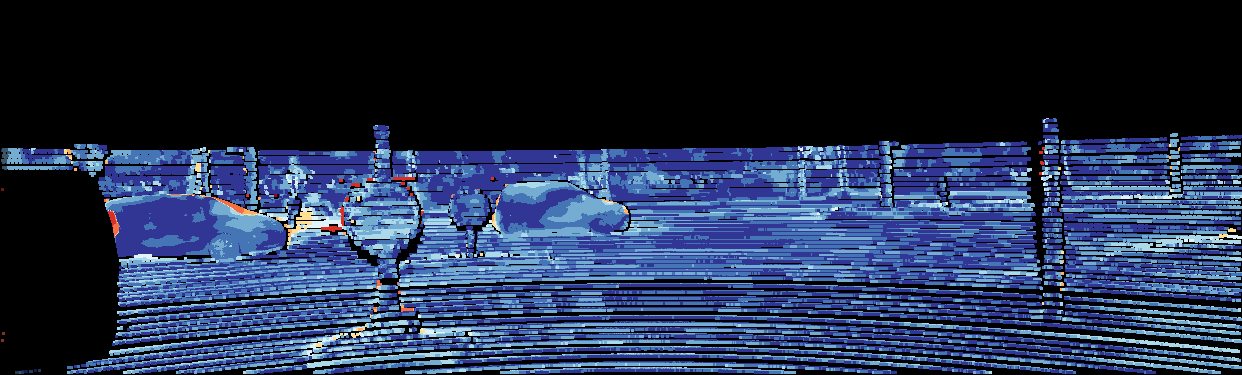} & \includegraphics[width=0.30\linewidth]{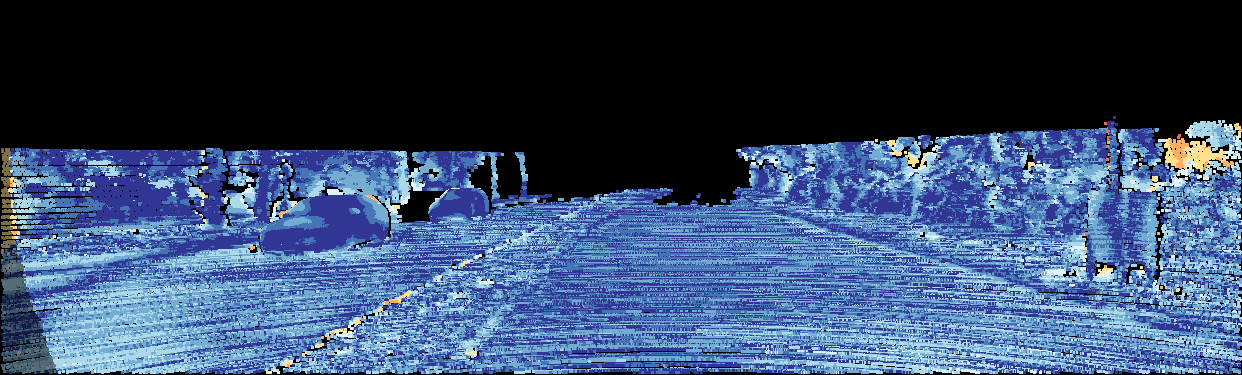} & \includegraphics[width=0.30\linewidth]{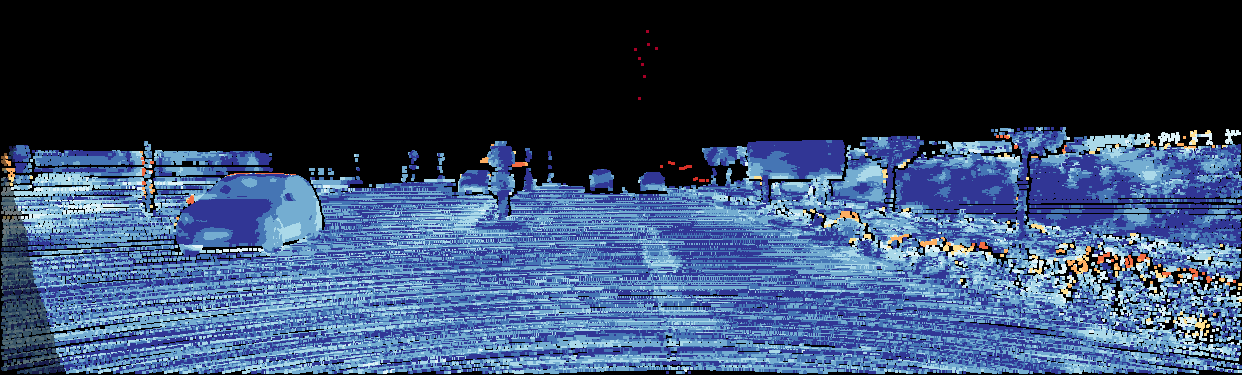} \\
				& GANet\cite{zhang_ga-net:_2019} & \\
				
%
				
		\includegraphics[width=0.30\linewidth]{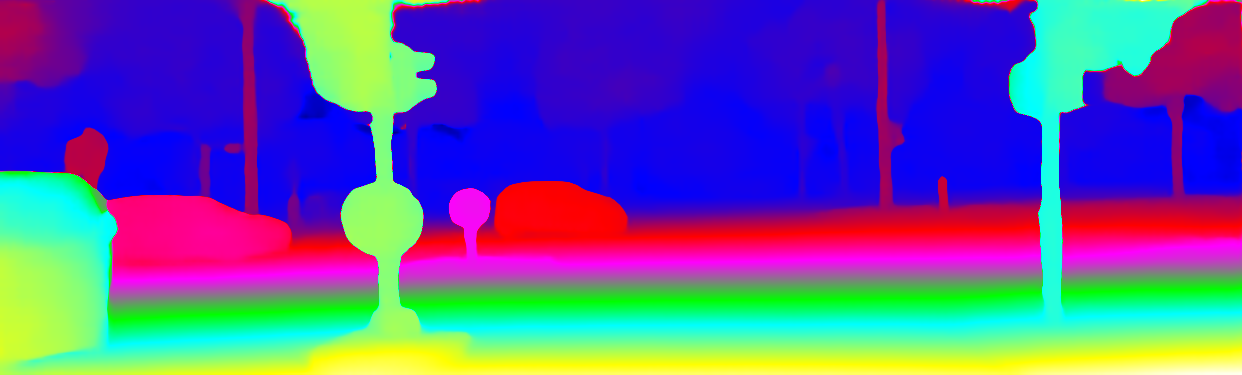} & \includegraphics[width=0.30\linewidth]{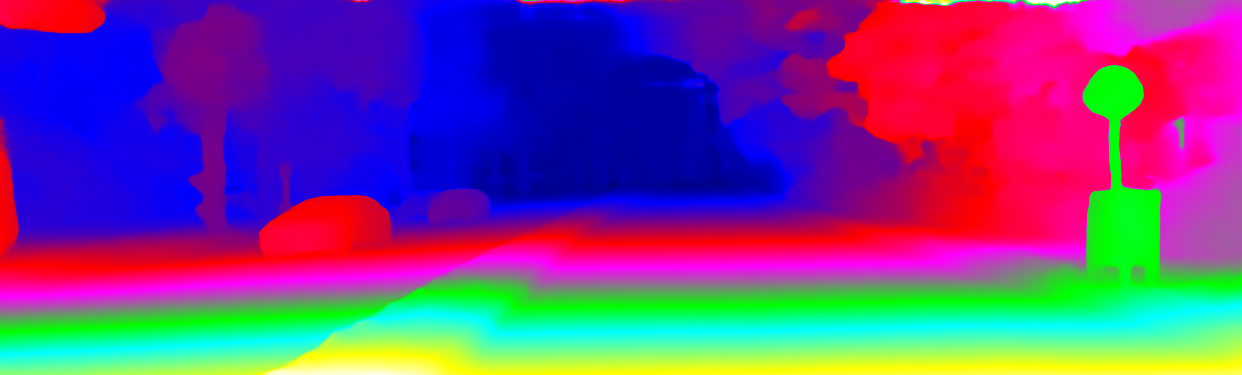} & \includegraphics[width=0.30\linewidth]{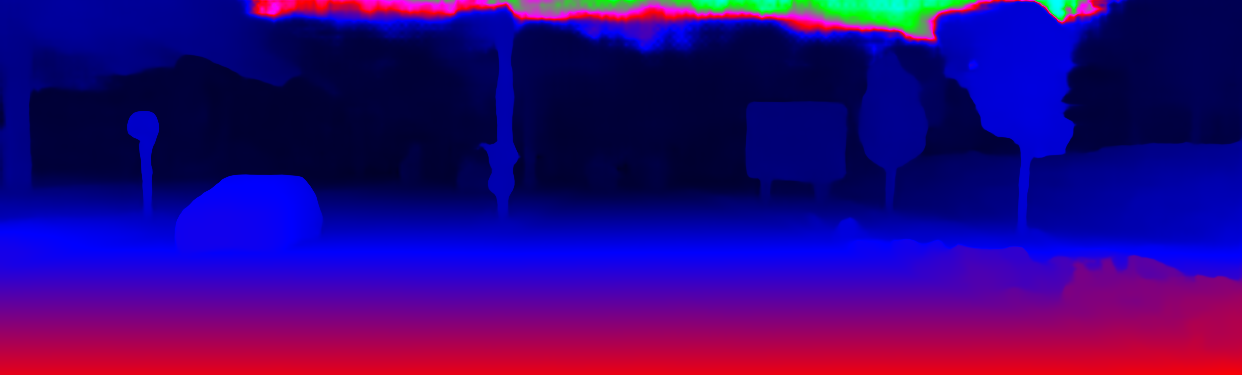} \\
		
		\includegraphics[width=0.30\linewidth]{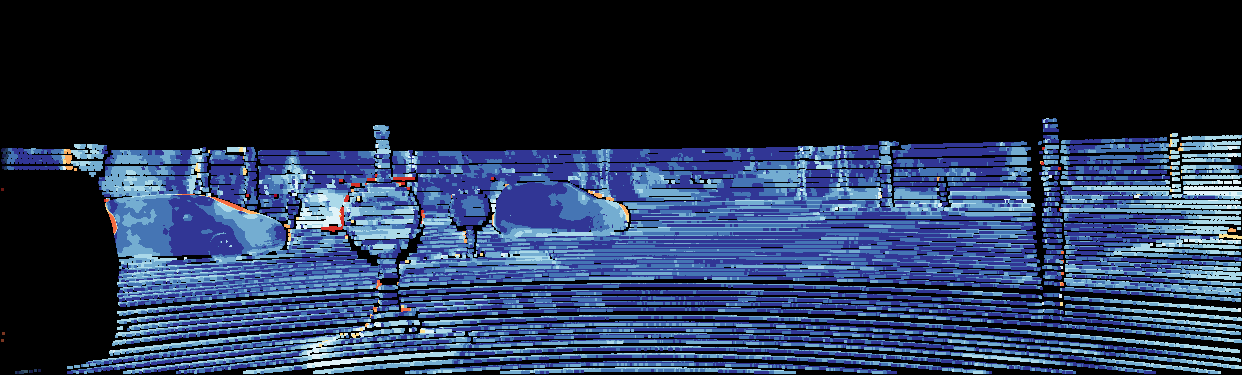} & \includegraphics[width=0.30\linewidth]{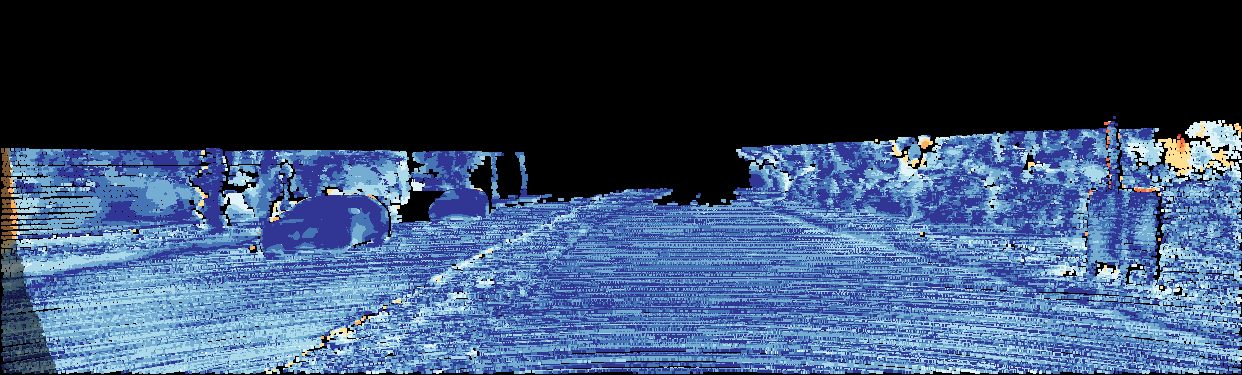} & \includegraphics[width=0.30\linewidth]{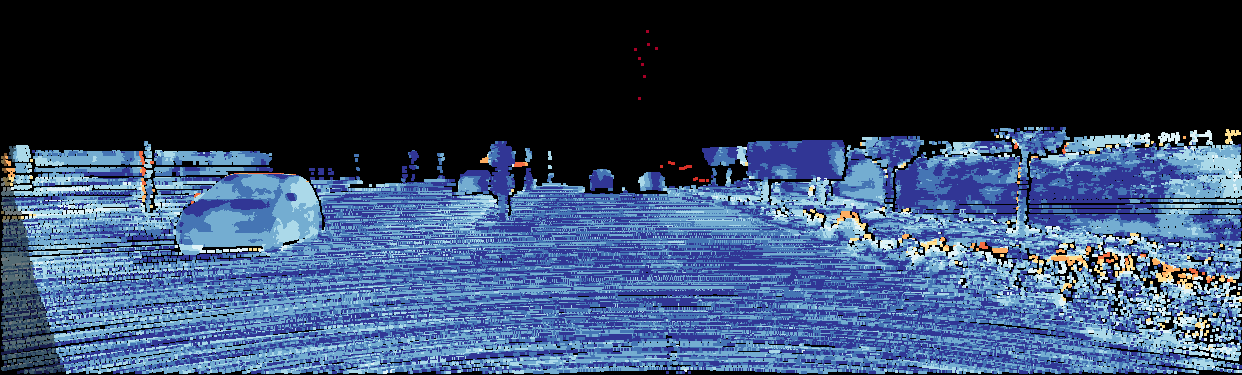} \\
		& GwcNet-g\cite{guo_group-wise_2019} & \\
		
		\includegraphics[width=0.30\linewidth]{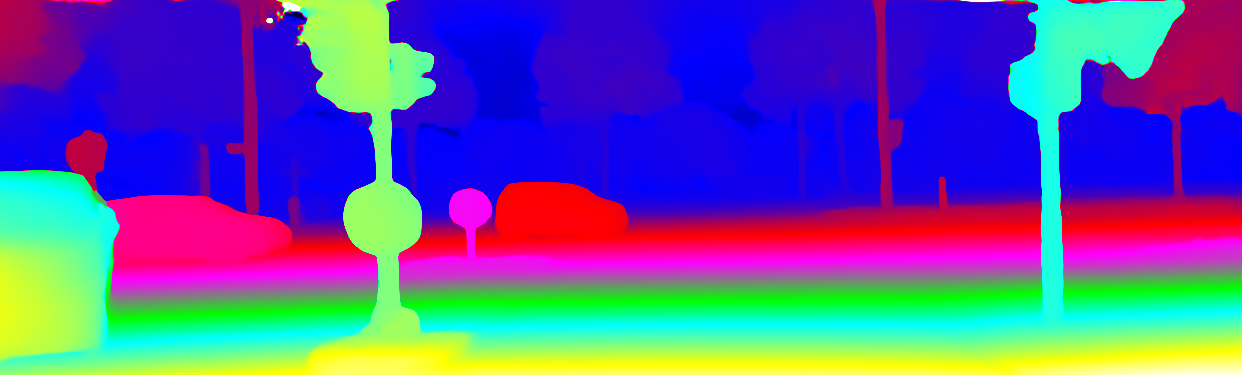} & \includegraphics[width=0.30\linewidth]{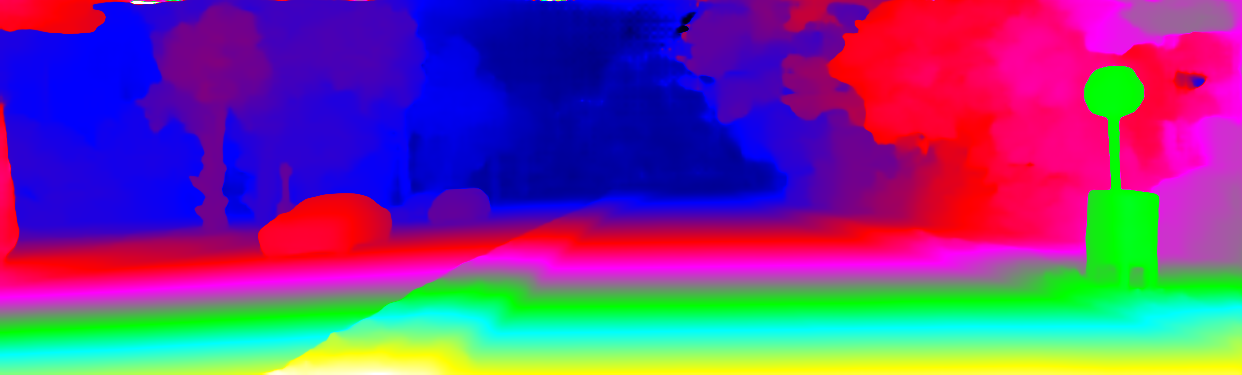} & \includegraphics[width=0.30\linewidth]{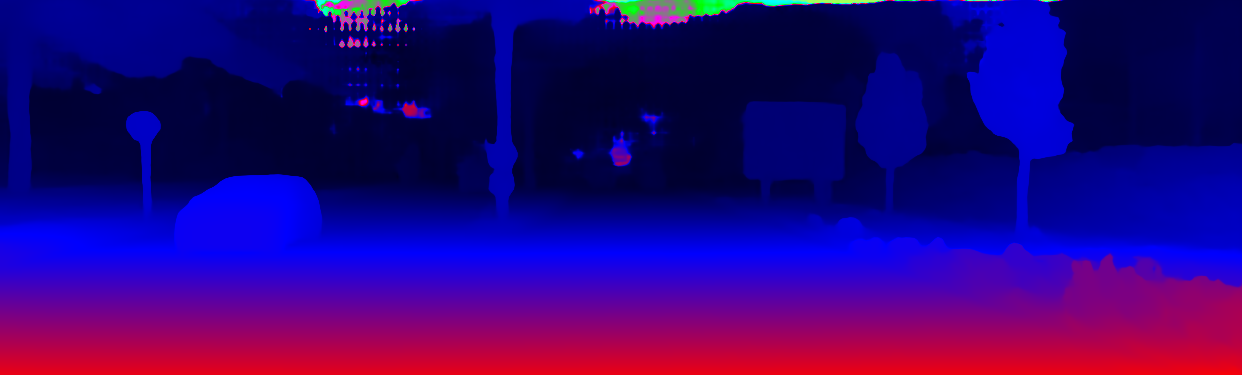} \\
		
		\includegraphics[width=0.30\linewidth]{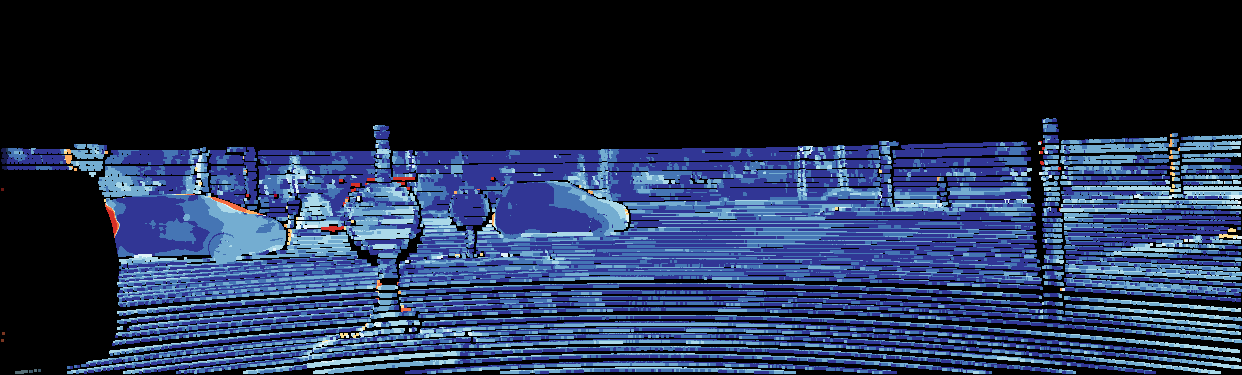} & \includegraphics[width=0.30\linewidth]{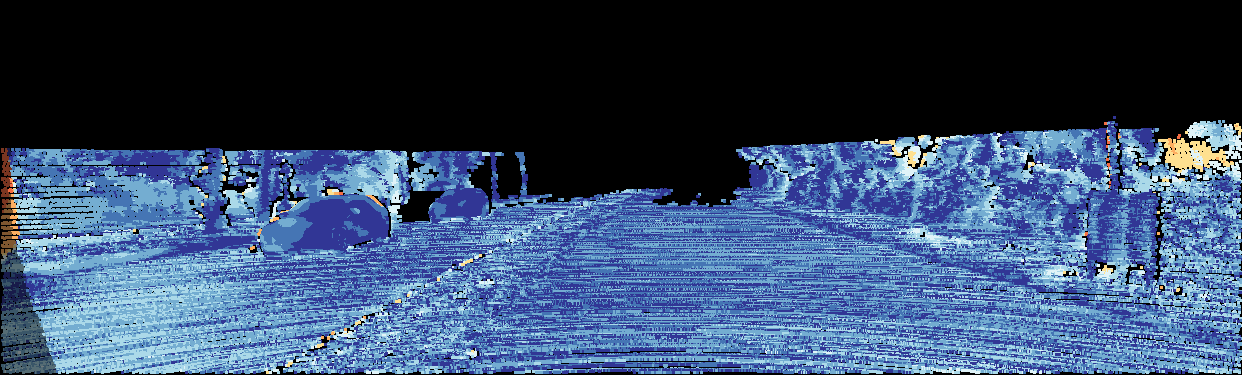} & \includegraphics[width=0.30\linewidth]{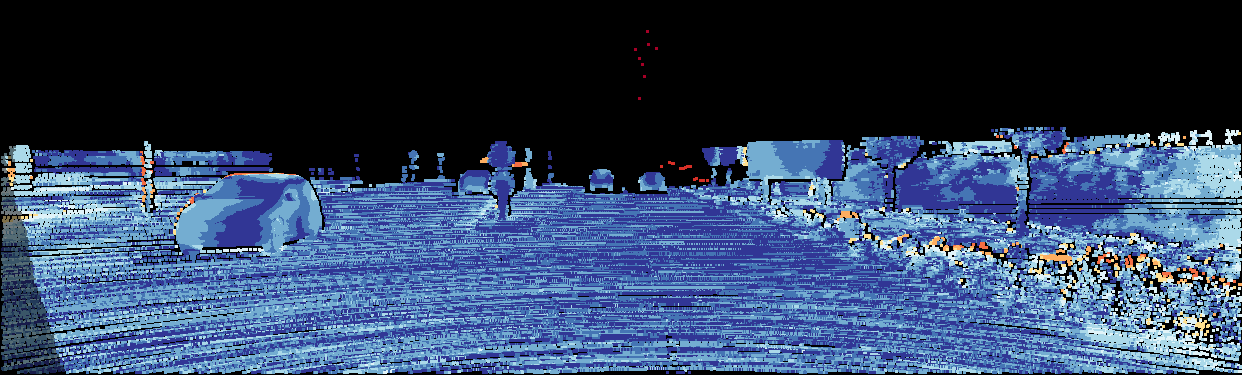} \\
		& ACVNet\cite{xu2022acvnet} & \\
		
		\includegraphics[width=0.30\linewidth]{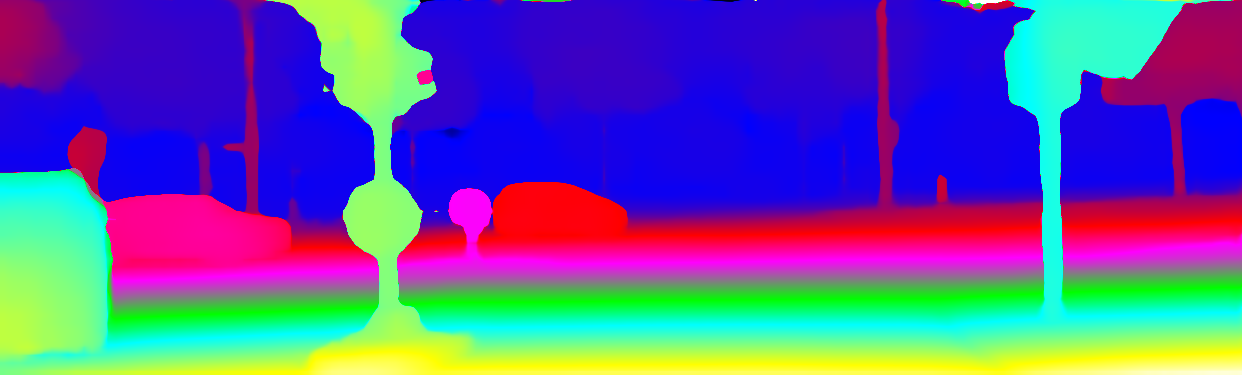} & \includegraphics[width=0.30\linewidth]{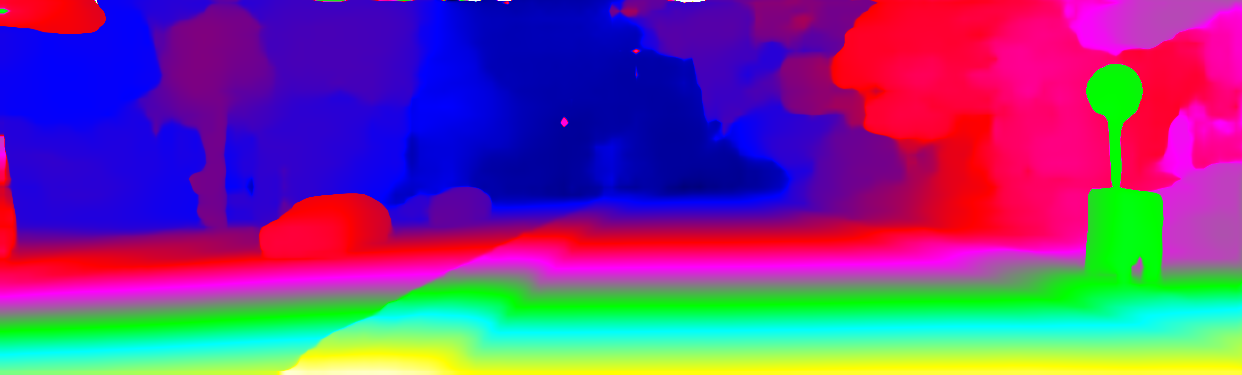} & \includegraphics[width=0.30\linewidth]{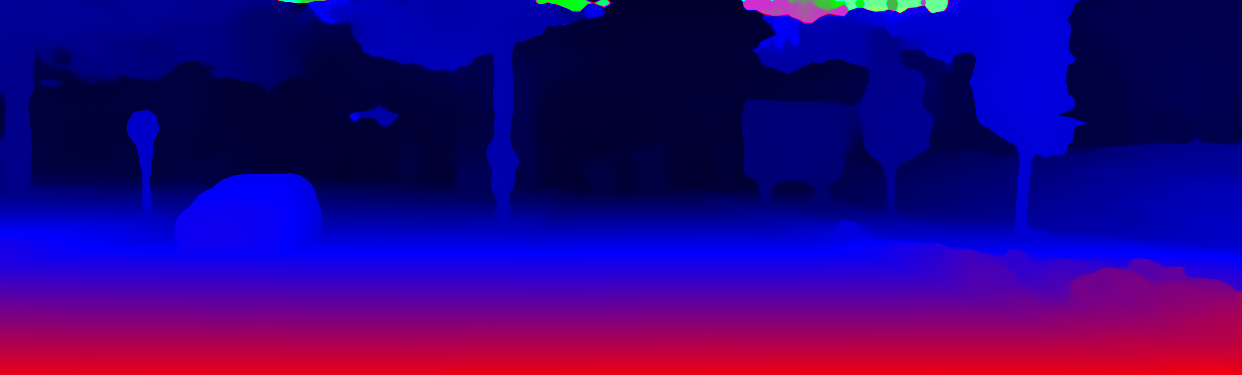} \\
		
		\includegraphics[width=0.30\linewidth]{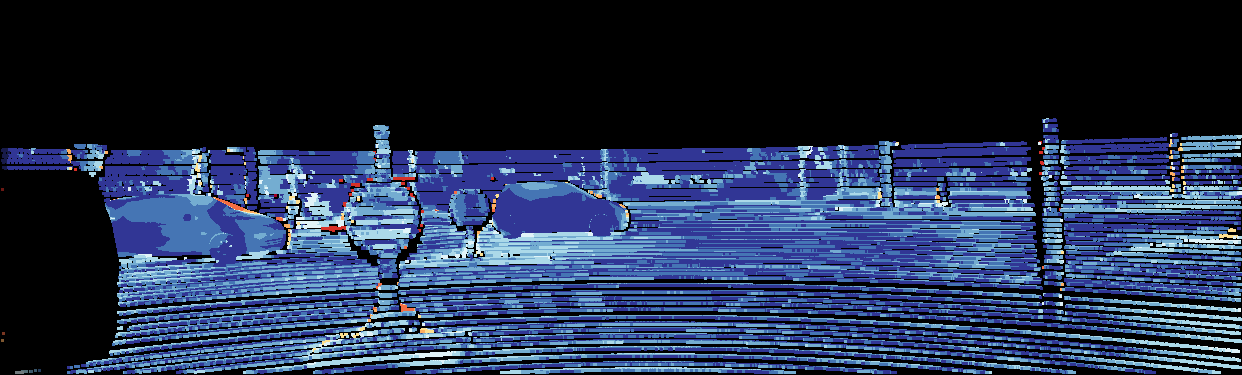} & \includegraphics[width=0.30\linewidth]{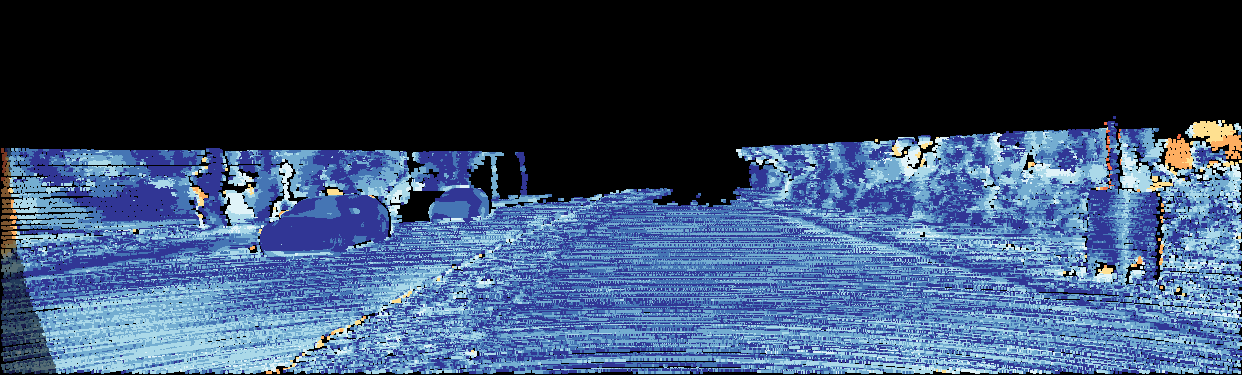} & \includegraphics[width=0.30\linewidth]{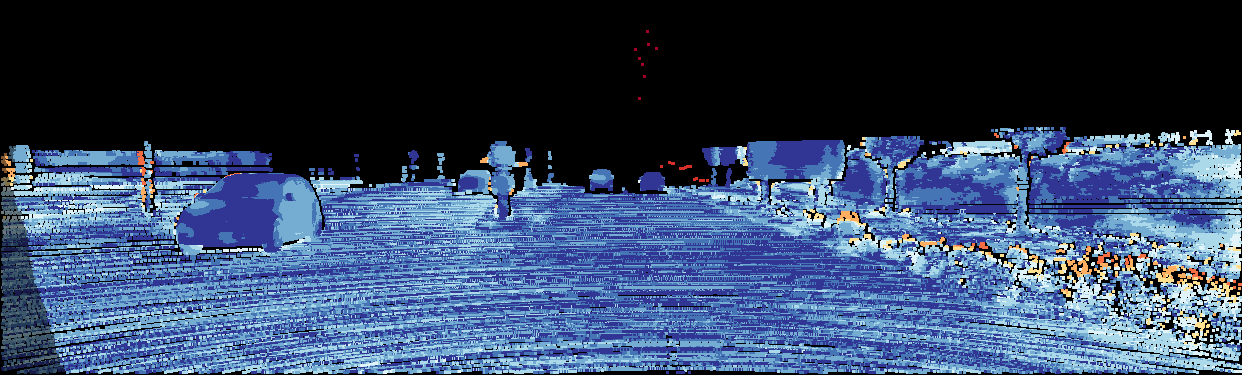} \\
		& \methodname  (Ours) & \\
	\end{tabular}
	\caption{Qualitative performance (disparity images together with error maps) from the \kittif benchmark. Warmer colors in error maps represent higher disparity errors.}
	\label{fig:kitti-benchmark-quality}
		\vspace*{-15pt}
\end{figure*}

%% file: results/memorytime.tex

\begin{table}[tbp]
	\centering
		\begin{adjustbox}{width=\columnwidth}
		\begin{tabular}{cc}
		\begin{tabular}{l|c|c}
			\hline				
			{Method} & EPE(px)& \makecell{Avg. Inference \\ time(msec)}\\				
			\hline
			PSMNet\cite{chang_pyramid_2018}  & 0.88 & 619 \\
			GA-Net-deep\cite{zhang_ga-net:_2019} & 0.84 & 3580 \\
			GA-Net-11\cite{zhang_ga-net:_2019}  & 0.93 & 2362\\			
			GwcNet-g\cite{guo_group-wise_2019}   & 0.79 & 240\\
			CFNet\cite{shen2021cfnet} & 0.97 & 483  \\
			LEAStereo\cite{cheng2020hierarchical} &0.78  & 945 \\
			ACVNet\cite{xu2022acvnet}  & 0.49 & 302 \\
			\hline
			\methodname (Ours) & 0.75 & 35 \tablefootnote{16bit Floating TensorRT model.}\\
			\hline
			\end{tabular}
				\begin{tabular}{l|c|c}
				\hline				
				{Method} & EPE(px)& \makecell{Avg. Inference \\ time(msec)}\\				
				\hline
			DispNetC\cite{mayer_large_2016} &1.68 & 60\\
			StereoNet\cite{khamis2018stereonet} & 1.1 & 15 \\
			DeepPruner-Fast\cite{duggal_deeppruner:_2019} & 0.97 & 61\\ 
			AANet\cite{xu_aanet_2020} & 0.87 & 62 \\
			DecNet\cite{yao2021decomposition} & 0.84 & 50\\
			HITNet\cite{tankovich2021hitnet} & 0.43 & 54\\
			\acvnet-Fast\cite{xu2022acvnet} & 0.77 &48\\
			EdgeStereo\cite{song_edgestereo_2018} & 0.74 & 320 \\

			\hline				
		\end{tabular}
			\end{tabular}
	\end{adjustbox}	

	\caption{ Inference time comparison of different 2D, real-time and \td Stereo methods.  The left table contains methods whose inference time have been recomputed using a uniform setting (see text) whereas timings for the methods in the right table are copied from the original publications.}
	\label{table:memory-time}
	\vspace*{-12pt}
\end{table}

%

%% file: LeanStereo.bbl
\begin{thebibliography}{10}
\providecommand{\url}[1]{#1}
\csname url@samestyle\endcsname
\providecommand{\newblock}{\relax}
\providecommand{\bibinfo}[2]{#2}
\providecommand{\BIBentrySTDinterwordspacing}{\spaceskip=0pt\relax}
\providecommand{\BIBentryALTinterwordstretchfactor}{4}
\providecommand{\BIBentryALTinterwordspacing}{\spaceskip=\fontdimen2\font plus
\BIBentryALTinterwordstretchfactor\fontdimen3\font minus
  \fontdimen4\font\relax}
\providecommand{\BIBforeignlanguage}[2]{{%
\expandafter\ifx\csname l@#1\endcsname\relax
\typeout{** WARNING: IEEEtran.bst: No hyphenation pattern has been}%
\typeout{** loaded for the language `#1'. Using the pattern for}%
\typeout{** the default language instead.}%
\else
\language=\csname l@#1\endcsname
\fi
#2}}
\providecommand{\BIBdecl}{\relax}
\BIBdecl

\bibitem{xu2022acvnet}
G.~Xu, J.~Cheng, P.~Guo, and X.~Yang, ``Acvnet: Attention concatenation volume
  for accurate and efficient stereo matching,'' \emph{arXiv preprint
  arXiv:2203.02146}, 2022.

\bibitem{cheng_learning_2019}
X.~Cheng, P.~Wang, and R.~Yang, ``\BIBforeignlanguage{en}{Learning {Depth} with
  {Convolutional} {Spatial} {Propagation} {Network}},''
  \emph{\BIBforeignlanguage{en}{arXiv:1810.02695 [cs]}}, Oct. 2019, arXiv:
  1810.02695.

\bibitem{shen2021cfnet}
Z.~Shen, Y.~Dai, and Z.~Rao, ``Cfnet: Cascade and fused cost volume for robust
  stereo matching,'' in \emph{CVPR}, 2021.

\bibitem{hosni2012fast}
A.~Hosni, C.~Rhemann, M.~Bleyer, C.~Rother, and M.~Gelautz, ``Fast cost-volume
  filtering for visual correspondence and beyond,'' \emph{IEEE TPAMI}, vol.~35,
  no.~2, 2012.

\bibitem{yoon2006adaptive}
K.-J. Yoon and I.~S. Kweon, ``Adaptive support-weight approach for
  correspondence search,'' \emph{IEEE TPAMI}, vol.~28, no.~4, 2006.

\bibitem{min2011revisit}
D.~Min, J.~Lu, and M.~N. Do, ``A revisit to cost aggregation in stereo
  matching: How far can we reduce its computational redundancy?'' in
  \emph{ICCV}.\hskip 1em plus 0.5em minus 0.4em\relax IEEE, 2011.

\bibitem{scharstein2002taxonomy}
D.~Scharstein and R.~Szeliski, ``A taxonomy and evaluation of dense two-frame
  stereo correspondence algorithms,'' \emph{ICCV}, vol.~47, no. 1-3, 2002.

\bibitem{poggi_synergies_2020}
M.~Poggi, F.~Tosi, K.~Batsos, P.~Mordohai, and S.~Mattoccia,
  ``\BIBforeignlanguage{en}{On the {Synergies} between {Machine} {Learning} and
  {Stereo}: a {Survey}},'' \emph{\BIBforeignlanguage{en}{PAMI}}, Apr. 2020.

\bibitem{hirschmuller_stereo_2008}
H.~Hirschmuller, ``\BIBforeignlanguage{en}{Stereo {Processing} by {Semiglobal}
  {Matching} and {Mutual} {Information}},'' \emph{\BIBforeignlanguage{en}{IEEE
  TPAMI}}, vol.~30, no.~2, Feb. 2008.

\bibitem{kendall2017endtoend}
A.~Kendall, H.~Martirosyan, S.~Dasgupta, P.~Henry, R.~Kennedy, A.~Bachrach, and
  A.~Bry, ``End-to-end learning of geometry and context for deep stereo
  regression,'' in \emph{ICCV}, 2017.

\bibitem{chang_pyramid_2018}
J.-R. Chang and Y.-S. Chen, ``\BIBforeignlanguage{en}{Pyramid {Stereo}
  {Matching} {Network}},'' in \emph{\BIBforeignlanguage{en}{CVPR}}.\hskip 1em
  plus 0.5em minus 0.4em\relax Salt Lake City, UT: IEEE, Jun. 2018.

\bibitem{zhang_ga-net:_2019}
F.~Zhang, V.~Prisacariu, R.~Yang, and P.~H.~S. Torr,
  ``\BIBforeignlanguage{en}{{GA}-{Net}: {Guided} {Aggregation} {Net} for
  {End}-to-end {Stereo} {Matching}},'' \emph{\BIBforeignlanguage{en}{CVPR}},
  Apr. 2019.

\bibitem{wang_anytime_2019}
Y.~Wang, Z.~Lai, G.~Huang, B.~H. Wang, L.~van~der Maaten, M.~Campbell, and
  K.~Q. Weinberger, ``\BIBforeignlanguage{en}{Anytime {Stereo} {Image} {Depth}
  {Estimation} on {Mobile} {Devices}},'' \emph{\BIBforeignlanguage{en}{ICCV}},
  Mar. 2019.

\bibitem{khamis2018stereonet}
S.~Khamis, S.~Fanello, C.~Rhemann, A.~Kowdle, J.~Valentin, and S.~Izadi,
  ``Stereonet: Guided hierarchical refinement for real-time edge-aware depth
  prediction,'' in \emph{ECCV}, 2018.

\bibitem{rahim2021separable}
R.~Rahim, F.~Shamsafar, and A.~Zell, ``Separable convolutions for optimizing 3d
  stereo networks,'' in \emph{ICCV}.\hskip 1em plus 0.5em minus 0.4em\relax
  IEEE, 2021.

\bibitem{shamsafar2022mobilestereonet}
F.~Shamsafar, S.~Woerz, R.~Rahim, and A.~Zell, ``Mobilestereonet: Towards
  lightweight deep networks for stereo matching,'' in \emph{WACV}, 2022.

\bibitem{sandler2018mobilenetv2}
M.~Sandler, A.~Howard, M.~Zhu, A.~Zhmoginov, and L.-C. Chen, ``Mobilenetv2:
  Inverted residuals and linear bottlenecks,'' in \emph{CVPR}, 2018.

\bibitem{wang2018pelee}
R.~J. Wang, X.~Li, and C.~X. Ling, ``Pelee: A real-time object detection system
  on mobile devices,'' \emph{NeurIPS}, vol.~31, 2018.

\bibitem{yu2021bisenet}
C.~Yu, C.~Gao, J.~Wang, G.~Yu, C.~Shen, and N.~Sang, ``Bisenet v2: Bilateral
  network with guided aggregation for real-time semantic segmentation,''
  \emph{ICCV}, vol. 129, no.~11, 2021.

\bibitem{cheng2020hierarchical}
X.~Cheng, Y.~Zhong, M.~t. Harandi, Y.~Dai, X.~Chang, H.~Li, T.~Drummond, and
  Z.~Ge, ``Hierarchical neural architecture search for deep stereo matching,''
  \emph{NeurIPS}, vol.~33, 2020.

\bibitem{tan2019efficientnet}
M.~Tan and Q.~Le, ``Efficientnet: Rethinking model scaling for convolutional
  neural networks,'' in \emph{ICCV}.\hskip 1em plus 0.5em minus 0.4em\relax
  PMLR, 2019.

\bibitem{mayer_large_2016}
N.~Mayer, E.~Ilg, P.~Hausser, P.~Fischer, D.~Cremers, A.~Dosovitskiy, and
  T.~Brox, ``\BIBforeignlanguage{en}{A {Large} {Dataset} to {Train}
  {Convolutional} {Networks} for {Disparity}, {Optical} {Flow}, and {Scene}
  {Flow} {Estimation}},'' in \emph{\BIBforeignlanguage{en}{CVPR}}.\hskip 1em
  plus 0.5em minus 0.4em\relax Las Vegas, NV, USA: IEEE, Jun. 2016.

\bibitem{liang_learning_2018}
Z.~Liang, Y.~Feng, Y.~Guo, H.~Liu, W.~Chen, L.~Qiao, L.~Zhou, and J.~Zhang,
  ``\BIBforeignlanguage{en}{Learning for {Disparity} {Estimation} through
  {Feature} {Constancy}},'' \emph{\BIBforeignlanguage{en}{CVPR}}, Mar. 2018,
  arXiv: 1712.01039.

\bibitem{tonioni2019real}
A.~Tonioni, F.~Tosi, M.~Poggi, S.~Mattoccia, and L.~D. Stefano, ``Real-time
  self-adaptive deep stereo,'' in \emph{CVPR}, 2019.

\bibitem{yang_segstereo_2018}
G.~Yang, H.~Zhao, J.~Shi, Z.~Deng, and J.~Jia, ``{SegStereo}: {Exploiting}
  {Semantic} {Information} for {Disparity} {Estimation},'' \emph{ECCV}, Jul.
  2018.

\bibitem{song_edgestereo_2018}
X.~Song, X.~Zhao, H.~Hu, and L.~Fang, ``\BIBforeignlanguage{en}{{EdgeStereo}:
  {A} {Context} {Integrated} {Residual} {Pyramid} {Network} for {Stereo}
  {Matching}},'' \emph{\BIBforeignlanguage{en}{ACCV}}, Sep. 2018.

\bibitem{kendall_end--end_2017}
A.~Kendall, H.~Martirosyan, S.~Dasgupta, P.~Henry, R.~Kennedy, A.~Bachrach, and
  A.~Bry, ``\BIBforeignlanguage{en}{End-to-{End} {Learning} of {Geometry} and
  {Context} for {Deep} {Stereo} {Regression}},''
  \emph{\BIBforeignlanguage{en}{ICCV}}, Mar. 2017, arXiv: 1703.04309.

\bibitem{guo_group-wise_2019}
X.~Guo, K.~Yang, W.~Yang, X.~Wang, and H.~Li,
  ``\BIBforeignlanguage{en}{Group-wise {Correlation} {Stereo} {Network}},''
  \emph{\BIBforeignlanguage{en}{CVPR}}, Mar. 2019.

\bibitem{duggal_deeppruner:_2019}
S.~Duggal, S.~Wang, W.-C. Ma, R.~Hu, and R.~Urtasun, ``Deeppruner: {Learning}
  {Efficient} {Stereo} {Matching} {Via} {Differentiable} {Patchmatch},''
  \emph{arXiv:1909.05845 [cs]}, Sep. 2019.

\bibitem{Hu2018}
J.~Hu, M.~Ozay, Y.~Zhang, and T.~Okatani, ``Revisiting single image depth
  estimation: Toward higher resolution maps with accurate object boundaries,''
  \emph{CoRR}, vol. abs/1803.08673, 2018.

\bibitem{watson2019}
J.~Watson, M.~Firman, G.~J. Brostow, and D.~Turmukhambetov, ``Self-supervised
  monocular depth hints,'' \emph{CoRR}, 2019.

\bibitem{Jae-Han2018single}
J.-H. Lee, M.~Heo, K.-R. Kim, S.-E. Wei, and C.-S. Kim, ``Single-image depth
  estimation based on fourier domain analysis,'' in \emph{CVPR}, 2018.

\bibitem{menze_object_2015}
M.~Menze and A.~Geiger, ``\BIBforeignlanguage{en}{Object scene flow for
  autonomous vehicles},'' in \emph{\BIBforeignlanguage{en}{CVPR}}.\hskip 1em
  plus 0.5em minus 0.4em\relax Boston, MA, USA: IEEE, Jun. 2015.

\bibitem{geiger2012we}
A.~Geiger, P.~Lenz, and R.~Urtasun, ``Are we ready for autonomous driving? the
  kitti vision benchmark suite,'' in \emph{CVPR}.\hskip 1em plus 0.5em minus
  0.4em\relax IEEE, 2012.

\bibitem{he2016deep}
K.~He, X.~Zhang, S.~Ren, and J.~Sun, ``Deep residual learning for image
  recognition,'' in \emph{CVPR}, 2016.

\bibitem{kingma2014adam}
D.~P. Kingma and J.~Ba, ``Adam: A method for stochastic optimization,''
  \emph{arXiv preprint arXiv:1412.6980}, 2014.

\bibitem{loshchilov2017decoupled}
\BIBentryALTinterwordspacing
I.~Loshchilov and F.~Hutter, ``Fixing weight decay regularization in adam,''
  \emph{CoRR}, vol. abs/1711.05101, 2017. [Online]. Available:
  \url{http://arxiv.org/abs/1711.05101}
\BIBentrySTDinterwordspacing

\bibitem{zbontar_stereo_2016}
J.~Zbontar, Y.~LeCun \emph{et~al.}, ``Stereo {Matching} by {Training} a
  {Convolutional} {Neural} {Network} to {Compare} {Image} {Patches},''
  \emph{Journal of Machine Learning Research}, vol.~17, no.~1, 2016.

\bibitem{yee2020fast}
K.~Yee and A.~Chakrabarti, ``Fast deep stereo with 2d convolutional processing
  of cost signatures,'' in \emph{WACV}, 2020.

\bibitem{saikia_autodispnet:_2019}
T.~Saikia, Y.~Marrakchi, A.~Zela, F.~Hutter, and T.~Brox,
  ``\BIBforeignlanguage{en}{{AutoDispNet}: {Improving} {Disparity} {Estimation}
  {With} {AutoML}},'' \emph{\BIBforeignlanguage{en}{ICCV}}, Oct. 2019, arXiv:
  1905.07443.

\bibitem{xu_aanet_2020}
H.~Xu and J.~Zhang, ``{AANet}: {Adaptive} {Aggregation} {Network} for
  {Efficient} {Stereo} {Matching},'' \emph{arXiv:2004.09548 [cs]}, Apr. 2020.

\bibitem{yao2021decomposition}
C.~Yao, Y.~Jia, H.~Di, P.~Li, and Y.~Wu, ``A decomposition model for stereo
  matching,'' in \emph{CVPR}, 2021.

\bibitem{tankovich2021hitnet}
V.~Tankovich, C.~Hane, Y.~Zhang, A.~Kowdle, S.~Fanello, and S.~Bouaziz,
  ``Hitnet: Hierarchical iterative tile refinement network for real-time stereo
  matching,'' in \emph{CVPR}, 2021.

\end{thebibliography}
